# ERANet: Edge Replacement Augmentation for Semi-Supervised Meniscus Segmentation with Prototype Consistency Alignment and Conditional Self-Training

Siyue Li[1], Yongcheng Yao[1,2], Junru Zhong[1], Shutian Zhao[1], Fan Xiao[3], Tim-Yun Michael Ong[4], Ki-Wai Kevin Ho[5], James F. Griffith[6], Yudong Zhang[7], Shuihua Wang[8], Jin Hong[9,*], Weitian Chen[1,6,*]

[1] CU Lab for AI in Radiology (CLAIR), Department of Imaging and Interventional Radiology, The Chinese University of Hong Kong, Hong Kong, China.
[2] School of Informatics, University of Edinburgh, Edinburgh, Scotland, United Kingdom.
[3] Department of Radiology, Shanghai Sixth People's Hospital Affiliated to Shanghai Jiao Tong University School of Medicine, Shanghai, China.
[4] Department of Orthopaedics & Traumatology, The Chinese University of Hong Kong, Hong Kong, China.
[5] Department of Orthopaedics & Traumatology, The Chinese University of Hong Kong Medical Centre, Hong Kong, China.
[6] Department of Imaging and Interventional Radiology, The Chinese University of Hong Kong, Hong Kong, China.
[7] School of Computer Science and Engineering, Southeast University, Nanjing, China.
[8] Department of Biological Sciences, School of Science, Xi'an Jiaotong Liverpool University, Suzhou, China.
[9] School of Information Engineering, Nanchang University, Nanchang, China.

Email: siyueli@link.cuhk.edu.hk; yc.yao@ed.ac.uk; jrzhong@link.cuhk.edu.hk; zhaoshutian@sjtu.edu.cn; freidaxiao@gmail.com; michael.ong@cuhk.edu.hk; kevinkiwaiho@gmail.com; griffith@cuhk.edu.hk; yudongzhang@ieee.org; Shuihua.Wang@xjtlu.edu.cn; hongjin@ncu.edu.cn; wtchen@cuhk.edu.hk

* Correspondence should be addressed to Jin Hong & Weitian Chen

**Abstract**

The meniscus, a fibrocartilaginous structure within the knee joint, plays an essential role in joint stability and the prevention of knee osteoarthritis (OA). Accurate segmentation of the meniscus from magnetic resonance imaging (MRI) is crucial for early diagnosis and monitoring of OA progression. However, manual segmentation is labor-intensive, while automatic approaches face challenges due to variability in meniscal morphology, partial volume effects, and low tissue contrast.

To address these challenges, we propose ERANet, a semi-supervised framework that effectively leverages both labeled and unlabeled data through anatomically guided augmentation, consistency regularization, and iterative pseudo label refinement. Central to ERANet is edge replacement augmentation (ERA), a meniscus-specific augmentation strategy that introduces plausible morphological perturbations by modifying peripheral meniscal regions with context-aware background information. ERA is tailored to address the unique anatomical variability of meniscal structures. Alongside ERA, ERANet incorporates two generalizable learning modules: prototype consistency alignment (PCA), which enforces feature compactness via prototype-guided regularization, and conditional self-training (CST), which selectively incorporates reliable pseudo labels based on their temporal stability. The synergistic interaction among these modules enables ERANet to handle small, low-contrast anatomical structures with limited supervision.

We validated ERANet on 3D DESS and 3D FSE MRI sequences, demonstrating superior segmentation performance compared to state-of-the-art semi-supervised methods. ERANet maintains high accuracy even with minimal labeled data, and extensive ablation studies confirm the individual and combined benefits of ERA, PCA, and CST. Our results suggest that ERANet offers a robust and scalable solution for meniscus segmentation. Code is available at https://github.com/SiYueLi-MRIandAI/ERANet.

## 1. Introduction

Knee osteoarthritis (OA) is a common and debilitating condition characterized by pain, stiffness, and disability, particularly in older adults [1]. Early detection and intervention are essential for managing OA and mitigating its long-term effects [2-4]. The meniscus, a crescent-shaped fibrocartilaginous structure within the knee joint, plays a critical role in shock absorption and load distribution between the femur and tibia [5]. Meniscal degeneration or injury not only impairs joint function but is also strongly associated with the onset and progression of knee OA [6, 7]. Due to its key biomechanical functions, the meniscus is often one of the first knee structures to exhibit degeneration or injury during OA progression [8]. Consequently, precise segmentation and quantification of the meniscus are crucial for tracking its morphological and compositional changes over time, facilitating early diagnosis and monitoring of disease progression.

Magnetic Resonance Imaging (MRI) has become the modality of choice for assessing meniscal integrity, owing to its superior soft tissue contrast and multiplanar imaging capabilities [9]. MRI enables the visualization of subtle meniscal abnormalities and provides diagnostic information that is not obtainable with conventional radiography or computed tomography [10]. However, accurate meniscus segmentation from MRI remains challenging due to its small size, thin crescentic morphology, and low contrast with adjacent tissues, especially in sequences like Fast Spin Echo (FSE), where the meniscus appears hypointense. While deep learning (DL) methods have achieved promising results in meniscus segmentation [11-16], their performance typically relies on extensive manual pixel-wise annotations, which are labor-intensive and costly to acquire in clinical practice.

Semi-supervised learning offers a promising solution by utilizing a small amount of labeled data with a larger pool of unlabeled data [17]. Various semi-supervised learning strategies have been applied to semantic segmentation, including self-training (ST) [19–22], adversarial learning [23–25], and consistency regularization [26–28]. While these approaches have demonstrated effectiveness in natural images and large-organ segmentation tasks, their direct application to meniscus MRI often results in limited performance. The meniscus lacks strong structural regularity and exhibits high morphological variability between patients. Furthermore, its indistinct boundaries and small area relative to background increase the risk of class imbalance and label noise, especially in pseudo-label-driven pipelines.

ST frameworks iteratively refine predictions using high-confidence pseudo labels, but are prone to error accumulation, especially when false positives are misclassified as stable targets [18-21]. Regularization-based methods such as adversarial learning [22-24] and consistency training [25-28] attempt to improve model generalization by enforcing prediction invariance under input perturbations. However, most rely on generic augmentations that may not be anatomically meaningful in the context of meniscus structures, potentially weakening the consistency constraint and undermining training stability. To mitigate such limitations, recent work has explored prototypical segmentation, which aligns unlabeled pixel-level features with class-wise prototypes to improve semantic compactness and label discrimination [29]. While effective in many tasks, this approach can struggle in grayscale MRI with weak texture cues, where features of adjacent classes often overlap. Moreover, the deformable-DETR framework with multi-level feature fusion [30], underscores the effectiveness of integrating spatial adaptability and hierarchical feature representations in biomedical imaging tasks.

These task-specific limitations motivate the development of a tailored framework for semi-supervised meniscus segmentation. To this end, we propose a novel semi-supervised framework for meniscus segmentation from MR images, termed ERANet, which integrates three tightly coupled components within a mean teacher (MT) [31] architecture: edge replacement augmentation (ERA), prototype consistency alignment (PCA), and conditional self-training (CST). Among them, ERA is specifically designed for meniscus segmentation, introducing anatomically meaningful perturbations by simulating structural variations of the meniscus. It achieves this by replacing peripheral meniscus regions with context-aware background information from the same image. Additionally, PCA improves intra-class feature compactness by computing the similarity between class prototypes and feature maps from unlabeled data, thereby encouraging consistent semantic representations across spatial locations. CST further enhances model robustness by selectively incorporating reliable pseudo labels based on prediction stability over time, and progressively refines previously uncertain samples through staged retraining. While the backbone architecture of ERANet remains general and domain-agnostic, the targeted design of ERA and the synergistic interaction between all components enable the framework to effectively address key challenges in meniscus segmentation under limited supervision.

We validated ERANet on two datasets acquired using 3D double-echo steady-state (DESS) and 3D Fast/Turbo Spin Echo (FSE/TSE) sequences, respectively. Due to the short T2 relaxation time of the meniscus and different echo times used in these pulse sequences, the contrast between the meniscus and the surrounding tissues is highly different in DESS and FSE images, with the meniscus being visible in DESS but completely dark in FSE images. Such contrast discrepancy demonstrates the generalization capability of the proposed approach.

Our contributions are summarized as follows:

1. We propose ERANet, a semi-supervised meniscus segmentation framework that integrates a task-specific augmentation strategy ERA, PCA, and CST. This framework significantly improves the segmentation accuracy of meniscus tissues with limited labeled data and achieves superior performance compared to state-of-the-art semi-supervised methods on datasets acquired using two different MRI pulse sequences.

2. ERA is a novel augmentation method tailored to the anatomical characteristics of the meniscus, simulating plausible variations in morphology and inflammation by replacing edge regions with localized background context. This enhances the quality of data perturbation and benefits subsequent learning stages, particularly for medial meniscus structures which are prone to injury and deformation.

3. PCA encourages feature consistency and semantic compactness by aligning pixel-level features with learned class prototypes. This alignment enhances representation quality for unlabeled data, particularly when labeled data is sparse.

4. CST improves segmentation robustness by filtering pseudo labels based on temporal stability and gradually incorporating all unlabeled data into training. This mitigates error accumulation and ensures more reliable ST.

5. Our extensive ablation studies demonstrate the synergistic effects of ERA, PCA, and CST in achieving robust performance across varying annotation levels. These techniques ensure reliable segmentation, achieving state-of-the-art results on datasets acquired using different MRI pulse sequences, even with sparse labeled data.

## 2. Related Work

### 2.1 Meniscus Segmentation

Since the introduction of the UNet architecture [32], medical image segmentation, including meniscus segmentation, has seen significant advancements. Multiple studies [12, 13, 33-35] have applied DL approaches, often augmented with post-processing techniques such as conditional random fields and statistical shape models, to enhance segmentation precision. Notably, Norman et al. [13] demonstrated strong concordance between manual and automated segmentations in deriving T1ρ and T2 relaxation times of meniscus. Similarly, Tack et al. [33] reported moderate correlations in medial meniscal extrusion assessment, reflecting DL's clinical utility. Adversarial learning has also been widely explored [15, 36]. Generative adversarial networks (GANs) are particularly well-suited for improving sensitivity to both global and local features, a critical requirement for accurately segmenting the meniscus's thin and irregular structure. These frameworks often include object-aware attention mechanisms [36] and iterative quality refinement [15] to mitigate under-segmentation and class imbalance. For instance, Jeon et al. [15] used a conditional GAN framework with pre-localization modules to focus segmentation on meniscus and cartilage regions.

However, the success of these methods rely heavily on extensive labeled datasets, presenting significant challenges due to the labor-intensive and costly nature of manual annotation. To reduce annotation burden, transfer learning and weak supervision have been investigated. Byra et al. [37] trained a 2D attention U-Net initialized on natural images, while Panfilov et al. [38] proposed a semi-supervised approach using limited annotated data. Xie et al. [39] extended this further using masked autoencoders pre-trained in a self-supervised fashion and guided by sparse point or line annotations. Nevertheless, these methods still require either large-scale annotations or task-specific priors and are often evaluated on a single MRI sequence.

In contrast, our proposed ERANet achieves robust segmentation performance across two distinct MRI sequences, 3D DESS and FSE, by explicitly modeling the anatomical variability of the meniscus. Moreover, ERANet significantly reduces annotation requirements (as low as 10% of the data) without compromising segmentation quality, supporting its scalability in practical deployment.

## 2.2 Semi-supervised Segmentation

Semi-supervised segmentation has evolved rapidly, offering strategies to effectively use large amounts of unlabeled data. Broadly, these strategies include ST [18-21, 40-44], adversarial learning [22-24], and consistency regularization [25-27, 45, 46]. Among these, the MT framework [34] is widely adopted in medical imaging, encouraging consistency between student and teacher models under input perturbations. Building on MT, later works integrated auxiliary modules for better regularization. For instance, [47] proposed leaking perturbations within a GAN-augmented MT framework to stabilize training, while [26] employed dual decoders with cyclic pseudo-labeling for left atrium segmentation. CPS [48] applied cross-pseudo supervision with twin networks, achieving strong results in multiple anatomical targets [49, 50]. Bidirectional copy-paste augmentation [51] was introduced to bridge the distribution gap between labeled and unlabeled data by exchanging semantic content.

Despite their success, these consistency-based methods often rely on generic augmentations and architectural assumptions that implicitly favor large, well-defined anatomical structures. However, such assumptions do not align with the intrinsic properties of the meniscus. Its small size and variable shape limit the effectiveness of conventional perturbations, which may weaken consistency loss and lead to suboptimal pseudo label generation. It often exhibits blurred or incomplete boundaries in MRI. Moreover, common architectural designs such as multi-scale fusion [52] and deep decoders often assume strong contextual cues or edge definitions, which are frequently absent in meniscal regions.

To mitigate intra-class variation, prototype-based consistency methods have gained popularity [29, 53-55]. Cyclic and contrastive prototype strategies [29, 54] enhance representation compactness and inter-class separation. However, most of these frameworks overlook the instability of prototype predictions when pseudo labels are noisy or the anatomical boundaries are ambiguous like the meniscus. Meanwhile, ST-based approaches have re-emerged as a promising direction [40-43], especially when combined with strong perturbations or ensemble-based filtering. Yet, these methods are mostly developed for natural images and are rarely adapted to medical segmentation tasks with ambiguous structures or weak contrast. Pseudo label propagation in such settings is especially prone to confirmation bias, as unreliable predictions can be reinforced over iterations without targeted correction mechanisms.

To address these gaps, we propose a semi-supervised framework ERANet tailored for meniscus segmentation. Unlike generic pipelines, ERANet integrates three purpose-built modules: (1) a structural augmentation strategy ERA to simulate realistic meniscal variation, (2) a PCA to enhance feature robustness under ambiguous supervision, and (3) a CST mechanism to selectively incorporate temporally stable pseudo labels. These modules are designed to function synergistically to overcome the structural, contrast, and annotation challenges inherent to meniscus segmentation.

## 3. Method

### 3.1 The Semi-supervised Segmentation Framework

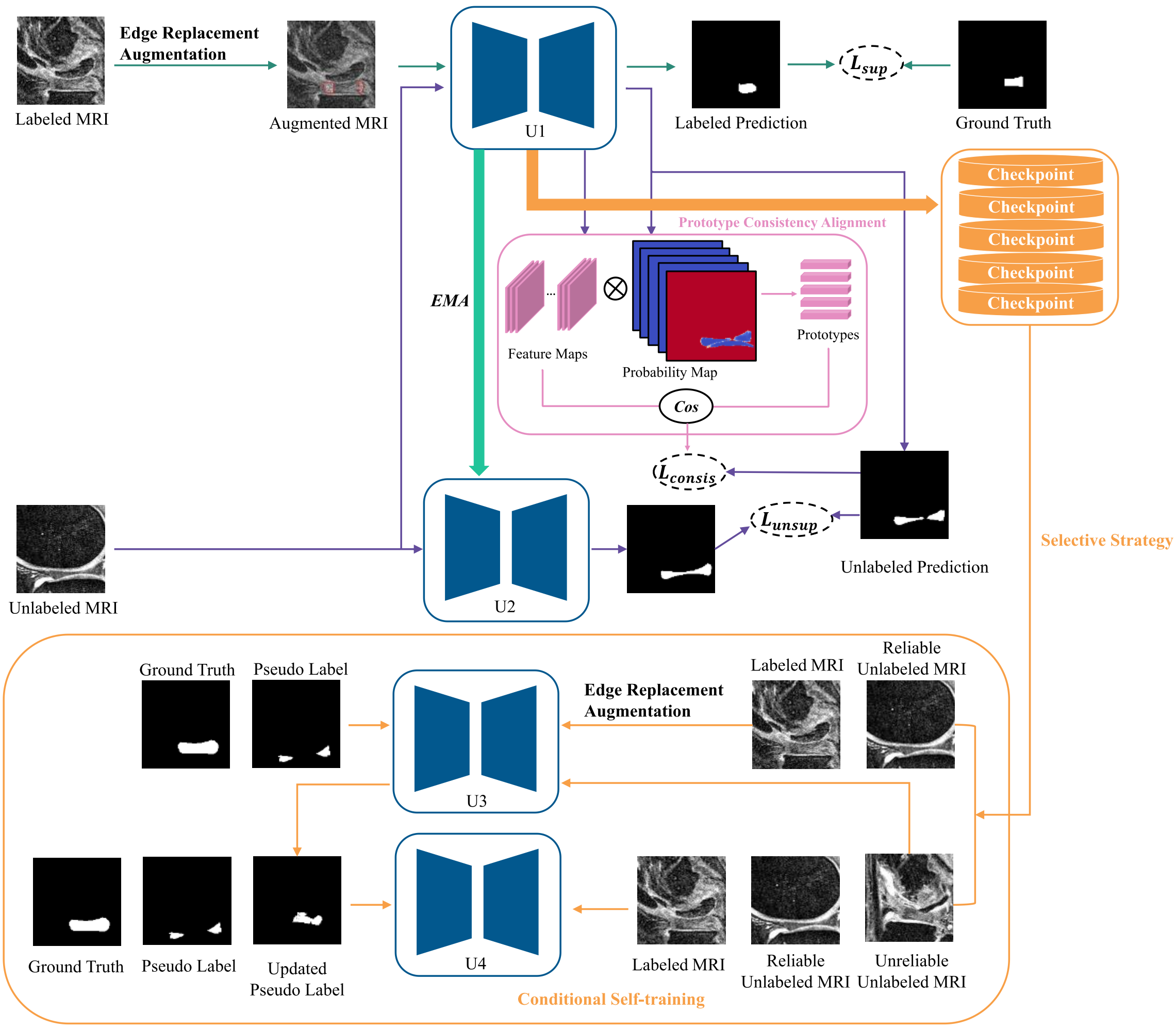

Figure 1. Architectural overview of ERANet, a semi-supervised meniscus segmentation framework leveraging both labeled and unlabeled images. ERANet comprises four segmentation networks (U1-U4), optimized through mean-teacher training, prototype consistency alignment, conditional self-training, and edge replacement augmentation. U1 is trained using a combination of supervised loss on labeled images, unsupervised loss on unlabeled images, and consistency loss derived from prototypical predictions on the unlabeled samples. U2 represents the Exponential Moving Average (EMA) of the U1 model. U3 and U4 are trained using both labeled and pseudo-labeled images. A selective strategy filters reliable unlabeled images, enabling effective conditional self-training and improving segmentation accuracy.

In this section, we introduce ERANet, a novel framework designed for semi-supervised meniscus segmentation, with its architectural overview depicted in Fig. 1. The framework leverages a labeled dataset $D_l = \{(x_i, y_i)\}_{i=1}^{n_l}$ and a significantly larger unlabeled dataset $D_u = \{x_i\}_{i=n_l+1}^{n_l+n_u}$ to train the proposed model. Central to the design are four standard segmentation networks (U1, U2, U3, and U4), which are systematically optimized using a combination of MT training, prototype consistency alignment, CST, and the innovative ERA technique. The ultimate goal is to develop a robust segmentation model (U4) capable of precise meniscus delineation from MR images. The training paradigm integrates both labeled and unlabeled data, achieving high performance with an annotation ratio as low as 10%. Notably, U2 is refined through an exponential moving average (EMA) of U1's weights. At inference, U4 directly outputs meniscus segmentations without requiring augmentation.

Three foundational innovations underpin the ERANet framework: (a) Edge Replacement Augmentation: ERA introduces structural variability by simulating meniscal shape and inflammatory variations. This is achieved by replacing the peripheral regions of the meniscus with background information (Fig. 1), thereby enhancing model robustness to morphological variations. (b) Prototype Consistency Alignment: This method strengthens intra-class feature similarity through prototype-based alignment. By computing correlations between prototypes and feature maps, it refines feature representations, while a prototypical consistency loss applied to unlabeled data fosters intra-class compactness. Cosine similarity serves as a pseudo-labeling mechanism to guide segmentation predictions. (c) Conditional Self-Training: This component advances segmentation accuracy by incorporating stable pseudo labels derived from U1's predictions across multiple checkpoints. Unlabeled images demonstrating consistent predictions are included in the labeled training set for U3, thereby iteratively enhancing model performance. Two rounds of ST are employed to maximize gains in segmentation precision.

As depicted in Fig. 1, U1 and U2 form the mean-teacher architecture, trained on both labeled and unlabeled images. ERA is applied as a specific augmentation technique for labeled images, while PCA is used to align feature representations from unlabeled images. During the training process, CST aids in selecting reliable unlabeled images. U3 is trained using all labeled data and the selected reliable unlabeled samples. It functions as an intermediate refinement step that improves pseudo label quality, particularly for previously uncertain regions. U4 is subsequently trained using the entire dataset, including all labeled images and the full set of unlabeled samples. For images not used in U3, U4 adopts the enhanced pseudo labels produced by U3, thereby incorporating both accurate supervision and comprehensive data coverage. This design ensures that the final segmentation model achieves a favorable balance between robustness and completeness. Using the proposed method, we concurrently perform tibial cartilage segmentation, as it is essential for downstream quantification tasks, including the calculation of meniscus thickness. Nevertheless, the performance evaluation is exclusively focused on meniscus segmentation metrics. Comprehensive methodological details and experimental results are provided in the following sections.

While each component of ERANet is introduced to address a specific challenge in semi-supervised segmentation, the full potential of the framework emerges from their integration. ERA focuses on data augmentation by introducing anatomically meaningful perturbations, PCA addresses feature alignment by promoting intra-class compactness, and CST improves pseudo label reliability through selective refinement. Although each module contributes individually, their combination yields a performance gain greater than the sum of their parts. ERA enhances feature diversity that benefits PCA, PCA improves prediction stability that strengthens CST, and CST enables progressive incorporation of unlabeled data with reduced noise. This synergy forms the foundation of ERANet's robustness and effectiveness, as confirmed by our ablation studies.

### 3.2 Edge Replacement Augmentation

We propose a novel data augmentation strategy, edge replacement augmentation, which is both conceptually simple and highly effective. The central objective of ERA is to simulate the variations in meniscal morphology and inflammation often observed in knee OA. This is achieved by replacing the peripheral regions of the meniscus with background information extracted from the same image. Unlike conventional augmentation methods such as Copy-Paste [56], which transfer image patches between distinct images, ERA implements intra-image augmentation. It achieves this by sampling pixel values from the background regions adjacent to the meniscus and generating random values within the estimated pixel intensity range of these regions to replace the marginal areas of the meniscus.

This strategy introduces structural variability into the training data, thereby enhancing the robustness of segmentation models to anatomical variations. In our ERANet framework, ERA is incorporated into the MT configuration for U1 training, creating a "weak-to-strong" augmentation pipeline that facilitates consistency learning. Additionally, ERA is applied as a strong perturbation during U3's ST process, ensuring further refinement of model predictions. Despite its simplicity, this approach yields substantial performance improvements when applied to models.

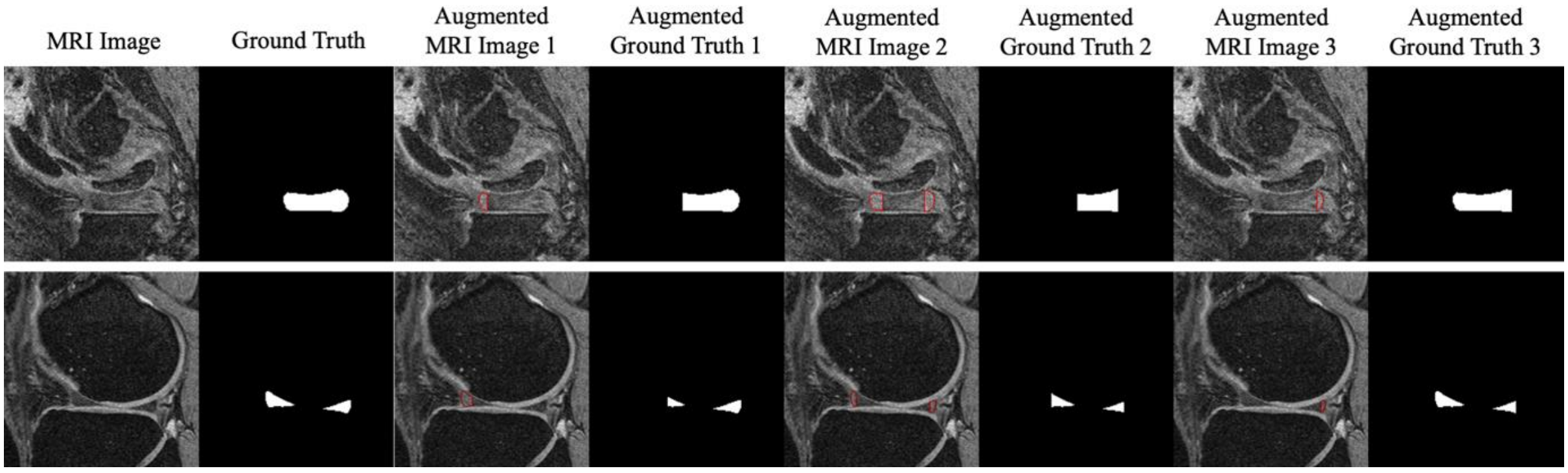

Figure 2. Representative examples of knee MRI images from the 3D DESS sequence alongside their edge-replaced augmented versions.

The example edge-replaced augmented images are presented in Fig. 2. To simulate anatomical variability along the meniscal margins, ERA replaces the edge regions, identified through anatomical extreme points, with background intensities sampled from adjacent non-meniscal tissue. This process emulates edge erosion or discontinuities commonly observed in challenging segmentation cases. ERA is implemented on labeled knee MR images and accounts for two anatomical configurations: (1) a single meniscal piece, referred to as *type A*, shown in the first row of Fig. 2, and (2) two distinct meniscal pieces, referred to as *type B*, depicted in the second row of Fig. 2. The augmentation process follows these steps:

1. **Mask Generation**
   Given a knee MRI image $X$ and its segmentation label $L$, a binary mask $M \in \{0, 1\}$ is created, where $M = 1$ denotes meniscus and $M = 0$ represents background. Subsequently, duplicate the image and mask to obtain $X_{ERA}$ and $M_{ERA}$, which will serve as the augmented outputs.
2. **Identification of Extreme Points**
   The extreme points along the outer curve of the meniscus in $M$ are identified as follows:
   - Far-left point ($P_{FL}$), upper-left point ($P_{UL}$), and bottom-left point ($P_{BL}$).
   - Far-right point ($P_{FR}$), upper-right point ($P_{UR}$), and bottom-right point ($P_{BR}$).

   For *type A* images (single meniscal piece), these points pertain to the single meniscus. For *type B* images (two separate meniscal pieces), $P_{FL}$, $P_{UL}$, $P_{BL}$ correspond to the left meniscal piece, and $P_{FL}$, $P_{UL}$, $P_{BL}$ to the right.
3. **Meniscus Distance and Replacement Point Selection**
   - The distance $d$ between $P_{FL}$ and $P_{FR}$ is measured for *type A* images. For *type B* images, the distances $d_L$ and $d_R$ are calculated for the respective pieces.
   - Replacement points $P_{RL}$ and $P_{RR}$ are randomly selected within a peripheral band along the outer meniscal margin, the extent of which is controlled by the edge-replacement ratio $r_e$. The ratio specifics the fraction of the meniscal width subjected to replacement:
     i. *type A* images:
        $P_{RL} \in (x_{FL}, x_{FL} + r_e d)$ and $P_{RR} \in (x_{FR} - r_e d, x_{FR})$
     ii. *type B* images:
        $P_{RL} \in (x_{FL}, x_{FL} + r_e d_L)$ and $P_{RR} \in (x_{FR} - r_e d_R, x_{FR})$

   where $x$ represents the $x$-coordinate of the point.

   The peripheral band is restricted to a moderate proportion of the meniscal width ($r_e = 0.3$ by default), ensuring that augmentation perturbs only peripheral structures while preserving the anatomical core. This empirically chosen range balances structural integrity and augmentation diversity. In sensitivity analyses, $r_e$ was varied to assess the robustness of the augmentation.
4. **Edge-Replaced Image Generation**
   - **Bounding Box Construction:**
     Two bounding boxes are defined to encompass the marginal meniscus regions.
     $BBox_L = ((x_{FL}\text{-}5, y_{BL}), (\max(x_{UL}, x_{BL}), y_{UL}))$
     $BBox_R = ((\min(x_{UR}, x_{BR}), y_{BR}), (x_{FR}\text{+}5, y_{UR}))$.
   - **Region Cropping:**
     Using these bounding boxes, the image $X$ and mask $M$ are cropped to produce $X_{CL}$, $X_{CR}$, $M_{CL}$, and $M_{CR}$, corresponding to the left and right regions of the mariginal meniscus, including surrounding tissues.

- **Background Creation:**
  Perform element-wise multiplication to remove the meniscus:
  $X_{CBL} = X_{CL} \times (1 - M_{CL})$ and $X_{CBR} = X_{CR} \times (1 - M_{CR})$, yielding background-only regions. The pixel value ranges $R_L$ and $R_R$ are then calculated for $X_{CBL}$ and $X_{CBR}$, respectively.
- **Pixel Replacement:**
  For pixels with x-coordinate $x < x_{RL}$, where $x_{RL}$ is the x-coordinate of $P_{RL}$, fill values using random samples from $R_L$ in $X_{ERA}$.
  For pixels with x-coordinate $x > x_{RR}$, where $x_{RR}$ is the x-coordinate of $P_{RR}$, fill values using random samples from $R_R$ in $X_{ERA}$.

5. **Edge-Replaced Mask Generation**
   For pixels with x-coordinate $x < x_{RL}$, where $x_{RL}$ is the x-coordinate of $P_{RL}$, set zero in $M_{ERA}$.
   - For pixels with x-coordinate $x > x_{RR}$, where $x_{RR}$ is the x-coordinate of $P_{RR}$, set zero in $M_{ERA}$.

$X_{ERA}$ and $M_{ERA}$ are the resulting augmented image and annotation respectively. Overall, ERA introduces targeted and anatomically plausible perturbations at meniscal boundaries.

### 3.3 Prototype Consistency Alignment

#### 3.3.1 Mean Teacher

The MT method is a semi-supervised learning approach that enhances segmentation by enforcing prediction consistency across augmented inputs while leveraging large unlabeled datasets [31]. It consists of a student network (U1) and a teacher network (U2) with identical architectures. In our ERANet framework, the student network is defined as $f_{U1}(X^l_{ERA}, Y, X^u; \theta)$, and the teacher network as $f_{U2}(X^u; \theta')$, where $f(\cdot)$ represents the segmentation network, and $\theta, \theta'$ are their parameters. The student is trained on both labelled ($X^l$) and unlabeled data $X^u$, with labeled examples augmented via ERA, while the teacher is updated through an EMA of the student's weights [31]. The teacher network provides pseudo labels for unlabeled data, guiding the student network toward robust predictions. The key mechanism, consistency regularization, minimizes discrepancies between student and teacher outputs on unlabeled data, improving segmentation accuracy and generalization.

#### 3.3.2 Prototypical Consistency Loss

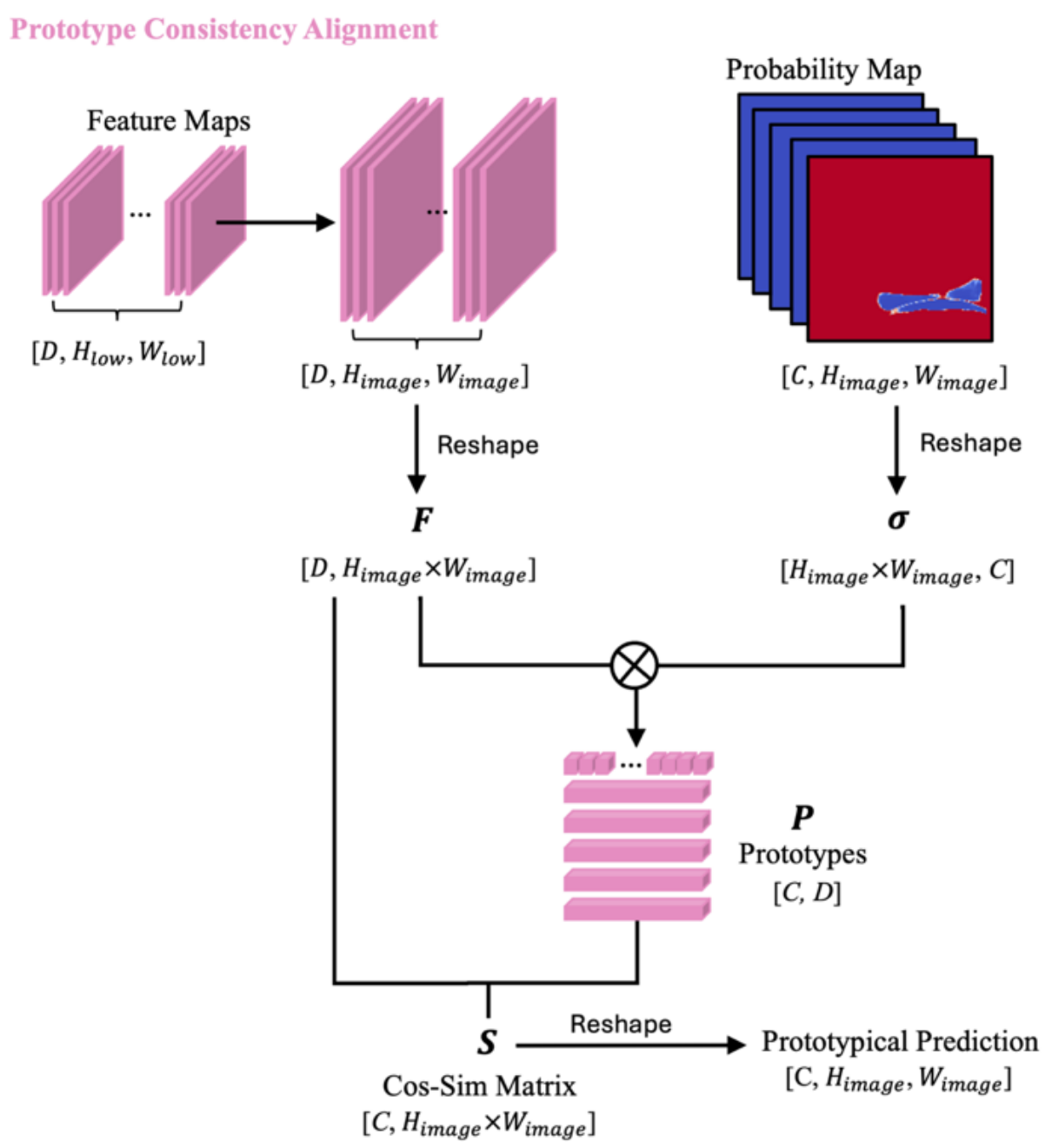

Figure 3. Detailed illustration of the prototypical consistency loss calculation.

Prototypes, as aggregated feature representations for each class, serve as condensed embeddings that encapsulate the defining characteristics of pixel-level features from input data. These prototypes act as class-specific feature centroids, guiding the segmentation network by leveraging the relationship between the prototypes and the input feature maps. Through the computation of correlations between prototypes and feature maps, the segmentation network reduces inter-class variability and enhances intra-class uniformity. In our framework, the similarity correlation maps are expressed as prototypical segmentation predictions, derived from the network's probability outputs and feature embeddings. This process is performed by U1 on unlabeled images, with the calculation steps for a single image illustrated in Fig. 3. To generalize the methodology, the figure demonstrates the application of prototypical predictions for both cartilage and meniscus segmentation. Formally, $\sigma$ denotes the output of the network. $F \in \mathbb{R}^{D \times H \times W}$ represents the feature map, where $D$ corresponds to the depth of the final up-convolutional block in the U-Net architecture. The prototypes for all classes is computed as:

$$P = \frac{F \cdot \sigma}{H \times W} \odot (1_D \cdot \frac{1}{\sum_{i=1}^{H \times W} \sigma_{i.}}) \tag{1}$$

This formulation ensures that the class prototype is calculated as a weighted average of feature embeddings, with weights determined by the network probabilities. By emphasizing regions with high confidence for class $j$, the prototype encapsulates the most discriminative characteristics of the class. This design not only enhances segmentation accuracy but also provides a robust mechanism for aligning feature spaces, which is critical for achieving high-quality results in semi-supervised settings. In each training iteration, U1 computes $B \times C$ prototypes, where $B$ denotes the batch size and $C$ represents the number of segmentation classes. To rigorously assess the alignment between the prototypes and the feature maps, the cosine similarity is calculated as:

$$S = \frac{P^T \cdot F}{||P||^T \cdot ||F||} \tag{2}$$

where $\frac{[\cdot]}{[\cdot]}$ denotes element-wise division, $||P|| \in \mathbb{R}^{\{1,C\}}$, and $||F|| \in \mathbb{R}^{\{1,H \times W\}}$. The norms $||P||_j$ and $||F||_j$ are defined as:

$$||P||_j \equiv \sqrt{\sum_{d=1}^{D} {P_{dj}}^2}\,, \; ||F||_j \equiv \sqrt{\sum_{d=1}^{D} {F_{dj}}^2}\,,$$

where $j$ indexes the dimensions of class $C$. Since the prototype is aggregated from the features within example itself, it ensures that $F$ aligns with more uniform feature representations, thereby promoting intra-class consistency in the segmentation predictions. This cosine similarity metric quantifies the angular alignment between the normalized feature embeddings and the corresponding class prototypes, reinforcing feature coherence and minimizing inter-class variance. To further promote intra-class cohesion and enhance segmentation accuracy, we propose a prototypical consistency loss to supervise the segmentation predictions.

$$L_{consis} = \sum_{j=1}^{C} \sum_{i}^{H \times W} (SS_{ij} - \sigma_{ij})^2 \tag{3}$$

where $SS_{ij}$ is defined as $SOFTMAX(S_{ij})$.

The incorporation of prototypical predictions supports the broader objective of minimizing reliance on labeled data by leveraging class-specific priors to refine segmentation outcomes, underscoring the potential of this methodology for advancing state-of-the-art segmentation techniques. Fig.4. illustrates the performance enhancement resulting from prototype consistency alignment.

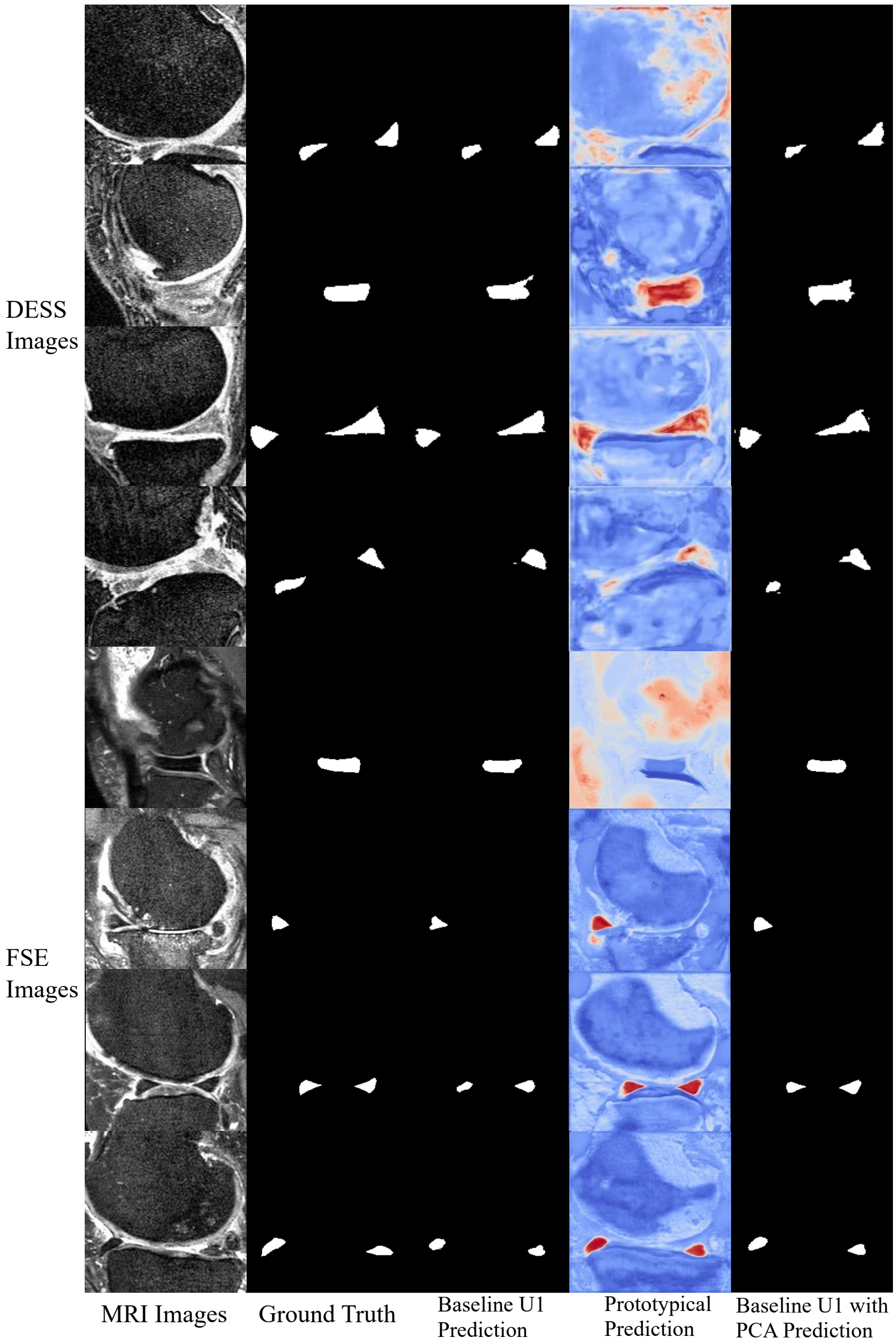

Figure 4. Representative examples of MRI images, including both DESS and FSE sequences, ground truth annotations of the meniscus, segmentation predictions from the baseline U1 model, and predictions from baseline U1 enhanced with PCA.

### 3.4 Conditional Self-Training Strategy

The modified MT framework, incorporating ERA and PCA, demonstrates strong performance. However, traditional ST methods fail to differentiate between unlabeled samples of varying stability and difficulty, treating all samples uniformly during training [40, 57]. This limitation arises from the absence of ground truth for unlabeled data, where predictions by the U1 may be inaccurate, leading to noisy pseudo labels that degrade performance when directly incorporated. To overcome this challenge, we propose a CST strategy that selectively includes pseudo labels based on their temporal reliability. Specifically, pseudo labels are only adopted for training if they demonstrate prediction stability across time, serving as the "condition" for inclusion. This filtering mechanism ensures that only high-confidence, consistent predictions are incorporated during the early stage of self-training, while harder examples are deferred until the model improves. Through two rounds of progressive

ST, CST enables the efficient and safe utilization of the entire unlabeled dataset, adopting an incremental learning strategy that incorporates increasingly complex samples over time.

Building upon insights from [40], we assess the reliability of pseudo labels for each unlabeled image $x_i \in D_u$ by leveraging their progressive stability during training. Specifically, for $K$ model checkpoints $\{M_j\}_{j=1}^{K}$ saved during the training process, we generate pseudo labels $\{O_{ij}^{U1}\}_{j=1}^{K}$ for $x_i$ using predictions from the student network *U1*. We systematically saved five checkpoints at equidistant intervals, corresponding to 1/5, 2/5, 3/5, 4/5, and 5/5 of the total training iterations. As models generally converge and achieve peak performance in the later stages of training, we evaluate the Dice Similarity Coefficient (DSC) between earlier pseudo labels and the final pseudo label to quantify prediction stability. The reliability score $R_i$ for the pseudo label of image $x_i$ is computed as follows:

$$R_i = \frac{1}{K-1}\sum_{j=1}^{K-1}\left(\frac{2O_{ij}^{U1}O_{iK}^{U1}}{O_{ij}^{U1}+O_{iK}^{U1}}\right) \tag{4}$$

where $R_i$ reflects the agreement between the intermediate pseudo labels $\{O_{ij}^{U1}\}_{j=1}^{K}$ and the final pseudo label $O_{iK}^{U1}$. A higher $R_i$ indicates greater stability and reliability.

Based on these scores, pseudo labels with $R_i \geq 0.8$ are selected for the first round of ST with model U3, while the rest are deferred. The threshold of 0.8 was empirically selected based on a sensitivity analysis (see Section 4.6.2), which demonstrated optimal performance around this value. After training U3, the model is employed to re-generate pseudo labels for the unreliable samples. The improvement in pseudo labeling for unreliable unlabeled images using U3 is illustrated in Fig. 5. In the second round, model U4 is trained using the full set of labelled and re-labeled data. The CST framework offers significant advantages by systematically incorporating stability metrics into the ST process. The pseudocode for the proposed CST method is detailed in Algorithm 1.

---

Algorithm 1: CST Pseudocode

---

**Input:** $D_l$ labeled training set, $D_u$ unlabeled training dataset, strong augmentation $ERA$, threshold $T$, $K$ checkpoints from U1 $\{M_j\}_{j=1}^{K}$
**Output:** Trained segmentation model U4
**for** $x_i \in D_u$ **do**
  **for** $M_j \in \{M_j\}_{j=1}^{K}$ **do**
    Predict pseudo label $O_{ij}^{U1} = M_j(x_i)$
    Compute reliability score $R_i$ for $x_i$ using Equation (4).
Select reliable samples $D_{sta} = \{x_k \mid R_k \geq T, x_k \in D_u\}$
Set $D_{unsta} = D_u \backslash D_{sta}$
Train U3 on $D_l \cup \{(x_k, M_K(x_k)) \mid x_k \in D_{sta}\}$ with ERA
Update pseudo labels for $D_{unsta}$ using *U3*
Train U4 on $D_l \cup \{(x_k, M_k(x_k)) \mid x_k \in D_{sta}\} \cup \{(x_k, \mathrm{U3}(x_k)) \mid x_k \in D_{unsta}\}$
**Return** U4

---

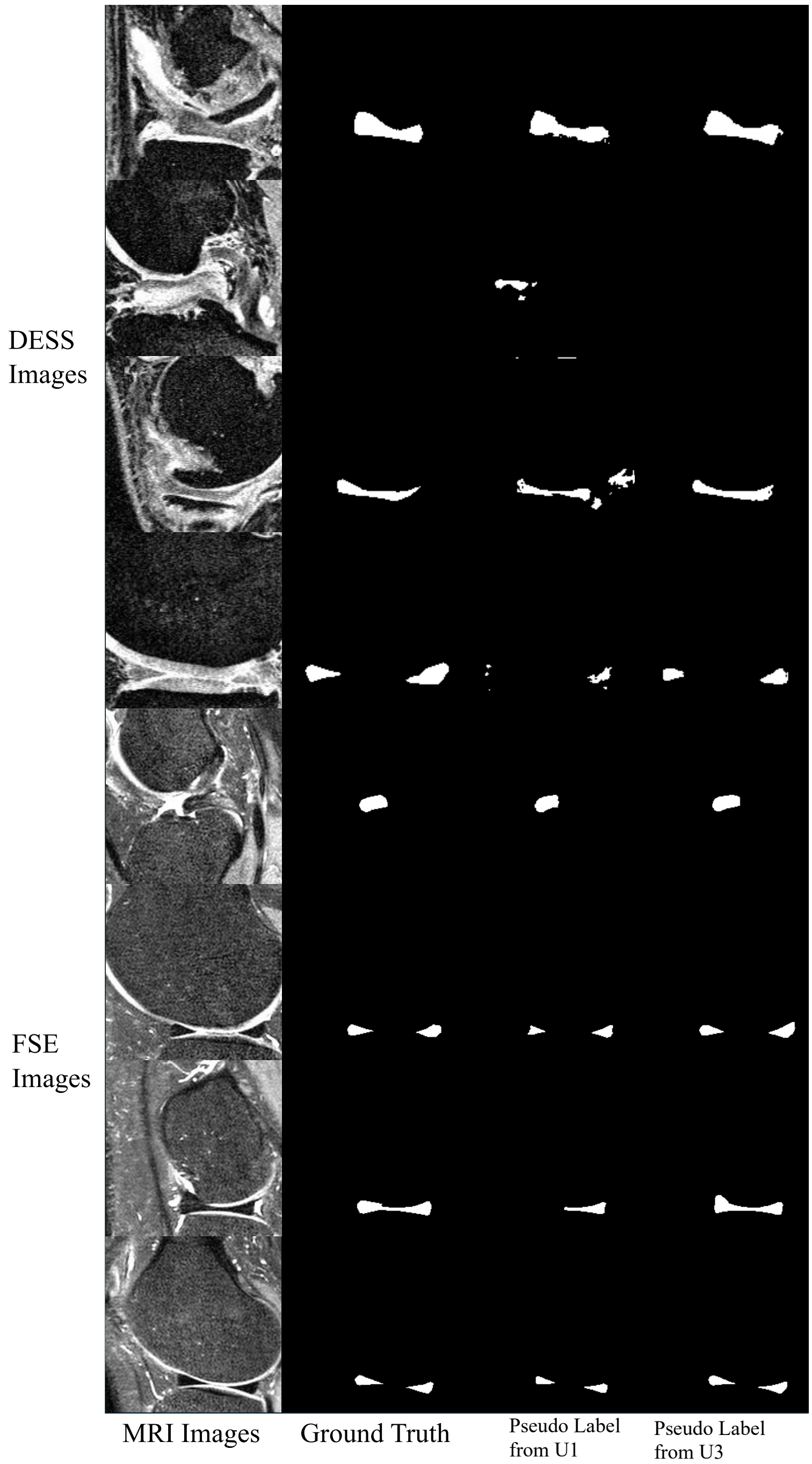

Figure 5. Representative examples of DESS and FSE MRI images, ground truth annotations for the meniscus, pseudo labels generated by the U1 model, and pseudo labels generated by the U3 model for unreliable unlabeled images.

### 3.5 Loss Function for Each Segmentation Network

To train the proposed semi-supervised segmentation framework (illustrated in Fig. 1), we define the loss functions for each network as follows. The loss function for the student network U1 is given by:

$$L_{U1} = \lambda_{sup} L_{sup}^{U1} + \lambda_{consis} L_{consis} + \lambda_{unsup} L_{unsup} \quad (5)$$

where $L_{sup}$ represents the supervised loss, which combines the DSC and cross-entropy loss for the labeled images. The prototypical consistency loss is weighted by a time-dependent factor $\lambda_{consis}$, defined as:

$$\lambda_{consis}(t) = 0.1 * e^{\left(-10\left(1-\frac{E}{E_{max}}\right)\right)^2} \quad (6)$$

where $E$ denotes the current epoch, and $E_{max}$ is the maximum epoch. $\lambda_{sup}$ and $\lambda_{unsup}$ are constant weights that control the relative importance of the supervised loss and the unsupervised loss, respectively. In our experiments, we empirically set $\lambda_{sup}$=1.0 and $\lambda_{unsup}$=1.0. The unsupervised loss $L_{unsup}$ measures the cross-entropy between the segmentation predictions from networks U2 and U1 on unlabeled image. For the segmentation network U3, the loss function $L_{U3}$ consists solely of the supervised loss, computed on the renewed labeled dataset. This dataset includes both the true labeled training data and the reliable pseudo-labeled data generated from previous models. For the segmentation network U4, the loss function $L_{U4}$ also consists solely of the supervised loss, calculated on the entire labeled dataset combined with the pseudo-labeled dataset. For reliable unlabeled images, pseudo labels are generated by U1, while for unreliable unlabeled images, updated pseudo labels are provided by U3. Both $L_{U3}$ and $L_{U4}$ utilize a combination of DSC and cross-entropy losses.

## 4. Experiments and Results

### 4.1 Datasets

We evaluated the performance of our semi-supervised segmentation models using two distinct datasets: the publicly available OAI-iMorphics dataset and an in-house dataset. The OAI-iMorphics dataset consists of 176 MR examinations, acquired using a 3.0T Siemens MAGNETOM Trio scanner (Siemens Healthcare, Erlangen, Germany) with a sagittal 3D DESS water excitation imaging sequence. This dataset is publicly accessible via the Osteoarthritis Initiative (OAI) repository at http://www.oai.ucsf.edu. The relevant imaging parameters for this dataset include a repetition time (TR) of 16.3 ms, an echo time (TE) of 4.7 ms, a flip angle of 25°, a voxel size of 0.365 × 0.465 × 0.7 mm, a matrix size of 307 × 384, a field of view of 140 mm, and 160 image slices. Further details can be found on the OAI website. Manual annotations for the knee joint regions were performed in the sagittal plane by two musculoskeletal radiologists, covering six distinct sub-regions: femoral cartilage, lateral tibial cartilage, medial tibial cartilage, patellar cartilage, lateral meniscus, and medial meniscus. For our experiments, we excluded the femoral and patellar cartilage regions, as they were not directly relevant to meniscus quantification. The dataset was randomly split into 140 subjects for training and 36 subjects for testing.

The in-house dataset consists of 33 knee MRI scans of patients with OA. The dataset was collected in 2021 at the Prince of Wales Hospital in Sha Tin, New Territories, Hong Kong SAR, China. These scans were acquired using a Philips Achieva TX 3.0T scanner (Philips Healthcare, Best, Netherlands) with a 3D FSE/TSE (VISTA™) sequence. The study was approved by the institutional review board. Among the 33 patients, 12 were classified as Kellgren-Lawrence (KL) grade 1, 9 as KL grade 2, 5 as KL grade 3, and 7 as KL grade 4. Severe OA cases (KL grade 3 and 4) make meniscus segmentation more challenging due to reduced contrast between meniscus and surrounding tissues. The imaging parameters for this dataset include a TR of 1200 ms, a TE of 32 ms, an acquisition voxel size of 0.55 × 0.545 × 0.55 mm and interpolated to a reconstruction voxel size of 0.29 × 0.29 × 0.55 mm, a matrix size of 236 × 276, and 292 image slices. The same six sub-regions (femoral cartilage, lateral tibial cartilage, medial tibial cartilage, patellar cartilage, lateral meniscus, and medial meniscus) were manually annotated by an experienced researcher and a radiologist with over 10 years of expertise. The dataset was randomly divided into 27 subjects for training and 6 subjects for testing.

In the context of our semi-supervised experiments, we treated a portion of the annotated training data as unannotated to simulate limited-label conditions. Prior to model training, we applied a series of preprocessing steps to both labeled and unlabeled images to improve image contrast and ensure consistency across datasets. Specifically, histogram equalization was used to enhance soft tissue contrast, particularly in FSE sequences where the meniscus appears dark. Intensity normalization was performed by scaling pixel values to the [0, 1] range based on the 1st and 99th percentile intensities. All images were then center-cropped to a standardized input size of 196 × 196 pixels. Additionally, N4 bias field correction was applied to the in-house dataset to mitigate low-frequency intensity inhomogeneity. These preprocessing steps were applied consistently across all experiments. To quantitatively evaluate segmentation performance, we adopted two widely used metrics in medical image segmentation: the DSC and the Average Symmetric Surface Distance (ASSD).

### 4.2 Implementation Details

The proposed semi-supervised segmentation framework was implemented in Python 3.9.7 with PyTorch 2.4.0. All experiments were conducted on two NVIDIA RTX A6000 GPUs, each equipped with 48 GB of memory, operating in parallel. The networks were initialized using Kaiming's normal initialization method. Consistent hyperparameters were maintained across the training of U1, U3, and U4. The Adam optimizer was used for training all networks, with an initial learning rate of 0.01. A polynomial decay schedule, defined as $lr = baselr(1 - \frac{epoch}{total_epoch})^{0.9}$, was employed to reduce the learning rate progressively. Each network was trained for 200 epochs with a batch size of 8, and these configurations were applied uniformly to both the OAI-iMorphics and the in-house datasets. To enhance the model's robustness against edge irregularities, strong data augmentations through ERA were applied to the unlabeled images during the training of U1 and U3. For computational efficiency, all images were cropped to a resolution of 256×256 pixels. Model training and evaluations were performed on a system powered by CUDA 11.5, enabling efficient accelerated computations.

### 4.3 Performances of the Proposed Semi-supervised Framework

Our proposed framework consists of four networks—U1, U2, U3, and U4—designed for meniscus segmentation from MR images, with the final segmentation results provided by the desired network, U4. Table 1 presents the quantitative evaluations of the segmentation performance of the proposed framework across the OAI-iMorphics and the in-house datasets. For the OAI-iMorphics dataset, the UNet model achieved optimal performance with a mean DSC of 88.95% when trained on the full set of labeled data. However, when the labeled data was reduced to 10% (n=14), the performance of baseline U1 model significantly declined, with the mean DSC dropping to 83.48%, highlighting the adverse effects of limited labeled data. The term "baseline U1" in the table refers to the UNet model trained exclusively on a small amount of labeled data, without the use of any unlabeled data or specialized training strategies. By leveraging 126 additional unlabeled scans through the proposed ERANet framework, the segmentation performance improved substantially, achieving a mean DSC of 86.66%. This represents a relative improvement of 3.8% over the baseline U1 model, demonstrating the effectiveness of incorporating unlabeled data.

For the in-house dataset, the UNet model trained on 27 labeled scans yielded a mean DSC of 73.12%. However, with only 3 labeled scans, the performance dropped dramatically, with the mean DSC falling to 50.38%. In this scenario, the integration of 24 unlabeled scans within the ERANet framework resulted in significant improvements, achieving a mean DSC of 64.33%. This corresponds to a relative improvement of 27.7% over the baseline U1 model trained with 3 labeled scans, highlighting the framework's ability to mitigate the challenges associated with limited labeled data, particularly in datasets characterized by greater complexity in imaging and anatomical structures. Across both datasets, the segmentation performance of the lateral meniscus consistently surpassed that of the medial meniscus, as evidenced by higher DSC values and lower ASSD values. These findings highlight the robustness and adaptability of the ERANet framework in handling diverse datasets and the scarcity of labeled data.

Table 1. Quantitative evaluation of segmentation performance of the UNet, Baseline U1, and the proposed framework. Baseline U1 refers to UNet model trained on a small amount of labelled data without the use of any unlabeled data.

| Dataset | Method | #scans used | | DSC (%) | | | ASSD (mm) | | |
|---|---|---|---|---|---|---|---|---|---|
| | | Labeled | Unlabeled | Lateral Meniscus | Medial Meniscus | Mean | Lateral Meniscus | Medial Meniscus | Mean |
| OAI-iMorphics | UNet | 140 | 0 | 90.63 | 87.26 | 88.95 | 0.14 | 0.20 | 0.17 |
| | Baseline U1 | 14 | 0 | 86.13 | 80.83 | 83.48 | 0.54 | 0.71 | 0.63 |
| | U4 (ERANet) | 14 | 126 | 88.86 | 84.46 | 86.66 | 0.22 | 0.43 | 0.33 |
| in-house | UNet | 27 | 0 | 82.47 | 63.76 | 73.12 | 0.51 | 2.23 | 1.37 |
| | Baseline U1 | 3 | 0 | 62.62 | 38.14 | 50.38 | 3.92 | 21.63 | 12.78 |
| | U4 (ERANet) | 3 | 24 | 78.14 | 50.51 | 64.33 | 1.31 | 7.39 | 4.35 |

### 4.4 Performance across Various Data Partitioning Strategies

This section evaluates the effectiveness of our proposed framework using a data partitioning approach, where the datasets are divided into labeled and unlabeled subsets with ratios of 1/n and (1-1/n), respectively. For the OAI-iMorphics dataset, the partitioning was applied with ratios of 1/20, 1/10, and 1/5. Due to the smaller size of the in-house dataset, the partitioning ratios used were 1/10, 1/5, and 1/2. Initially, we compared our approach with the fully supervised UNet, which was trained using the same labeled data ratios. Subsequently, we evaluated the performance of our proposed U1 and the final network, U4. U1 denotes the model enhanced with ERA and CST, building upon the baseline U1. The results of these evaluations for both datasets are summarized in Tables 2 and 3, respectively.

Table 2. Quantitative evaluation of the proposed method on the OAI-Imorphics dataset with different data partition protocols.

| Method | #scans used | | DSC (%) | | | ASSD (mm) | | |
|---|---|---|---|---|---|---|---|---|
| | Labeled | Unlabeled | Lateral Meniscus | Medial Meniscus | Mean | Lateral Meniscus | Medial Meniscus | Mean |
| Baseline U1 | 28 | 0 | 87.73 | 83.72 | 85.73 | 0.37 | **0.37** | 0.37 |
| U1 | 28 | 112 | 88.03 | 84.00 | 86.02 | 0.25 | 0.50 | 0.38 |
| U4 (ERANet) | | | **88.98** | **85.15** | **87.07** | **0.22** | 0.41 | **0.32** |
| Baseline U1 | 14 | 0 | 86.13 | 80.83 | 83.48 | 0.54 | 0.71 | 0.63 |
| U1 | 14 | 126 | 88.01 | 82.94 | 85.48 | **0.22** | 0.54 | 0.38 |
| U4 (ERANet) | | | **88.86** | **84.46** | **86.66** | **0.22** | **0.43** | **0.33** |
| Baseline U1 | 7 | 0 | 83.29 | 76.82 | 80.01 | 0.59 | 1.36 | 0.98 |
| U1 | 7 | 133 | 86.59 | 81.49 | 84.04 | 0.31 | 0.79 | 0.55 |
| U4 (ERANet) | | | **87.52** | **83.28** | **85.40** | **0.22** | **0.35** | **0.29** |

As shown in Table 2 for the OAI-iMorphics dataset, when trained with 1/5 of the labeled scans, the U1 network, which incorporated 112 unlabeled scans, demonstrated a slight improvement, achieving a mean DSC of 86.02%, up from 85.73% achieved by the UNet. The proposed ERANet further enhanced segmentation performance, attaining the highest mean DSC of 87.07%. Reducing the labeled data to 1/10 of the scans led to a decline in the UNet's performance, with a mean DSC of 83.48%. By leveraging 126 unlabeled scans, the U1 network improved the mean DSC to 85.48%. However, ERANet surpassed both models, achieving a mean DSC of 86.66%, underscoring its ability to effectively utilize unlabeled data to compensate for the reduction in labeled data. In the most challenging scenario, where only 7 labeled scans were available, the UNet achieved a mean DSC of 80.01%. By integrating 133 unlabeled scans, the U1 network improved to a mean DSC of 84.04%. Remarkably, ERANet excelled under these conditions, achieving a mean DSC of 85.40%, demonstrating its strong performance and potential in scenarios with minimal labeled data.

As the availability of labeled data decreased, the performance of the UNet declined markedly, reflecting the inherent challenges of training with limited annotations. In contrast, ERANet consistently mitigated this decline by leveraging unlabeled data, delivering substantial performance gains even in low-label settings. For instance, with only 7 labeled scans, ERANet achieved a mean DSC of 85.40%, significantly outperforming the UNet. These results emphasize ERANet's effectiveness in addressing the limitations of sparse annotations for meniscus segmentation. Its capacity to utilize unlabeled data efficiently positions it as a promising solution for improving segmentation accuracy in resource-constrained medical imaging environments.

Table 3 illustrates the performance of the models on the in-house dataset under varying proportions of labeled and unlabeled data. With 14 labeled scans, UNet achieved a mean DSC of 62.68%. The addition of 13 unlabeled scans enhanced the performance, with U1 achieving a mean DSC of 69.00%. ERANet further improved the outcome, reaching a mean DSC of 72.17%. When the labeled data was reduced to 5 scans, UNet's mean DSC dropped to 59.77%. However, incorporating 22 unlabeled scans into U1 increased the mean DSC to 65.42%, with ERANet outperforming both models at 67.09%. In the most challenging scenario, with only 3 labeled scans, UNet achieved a mean DSC of 50.38%. The inclusion of 24 unlabeled scans elevated U1's performance to a mean DSC of 60.73%. ERANet again demonstrated superior results, achieving a mean DSC of 64.33%. These findings underscore ERANet's effectiveness in leveraging unlabeled data to improve segmentation accuracy, particularly in settings with limited labeled examples.

Table 3. Quantitative evaluation of the proposed method on the in-house dataset with different data partition protocols.

| Method | #scans used | | DSC (%) | | | ASSD (mm) | | |
|---|---|---|---|---|---|---|---|---|
| | Labeled | Unlabeled | Lateral Meniscus | Medial Meniscus | Mean | Lateral Meniscus | Medial Meniscus | Mean |
| Baseline U1 | 14 | 0 | 75.81 | 49.54 | 62.68 | 1.62 | 6.26 | 3.94 |
| U1 | 14 | 13 | 80.75 | 57.26 | 69.01 | **0.66** | 4.15 | 2.41 |
| U4 (ERANet) | | | **81.08** | **63.25** | **72.17** | 0.79 | **2.09** | **1.44** |
| Baseline U1 | 5 | 0 | 72.00 | 47.53 | 59.77 | 1.51 | 6.17 | 3.84 |
| U1 | 5 | 22 | **78.64** | 52.19 | 65.42 | **0.73** | 5.69 | **3.21** |
| U4 (ERANet) | | | 78.17 | **56.01** | **67.09** | 1.06 | **5.66** | 3.36 |
| Baseline U1 | 3 | 0 | 62.62 | 38.14 | 50.38 | 3.92 | 21.63 | 12.78 |
| U1 | 3 | 24 | 73.49 | 47.96 | 60.73 | 2.59 | 10.6 | 6.60 |
| U4 (ERANet) | | | **78.14** | **50.51** | **64.33** | **1.31** | **7.39** | **4.35** |

## 4.5 Segmentation Performances Comparison with other Stage-of-the-arts Semi-supervised Methods

In this section, we compare our proposed semi-supervised segmentation framework with several state-of-the-art methods in the field. These methods include MT [31], UA-MT [58], AdvChain [59], CTCT [60], CLD [61], SCP-Net [55], DAE-MT [62], Hybrid-Teacher [63], LeFeD [46], and PICK [44]. Among these, UA-MT is one of the most widely used MT-based frameworks for semi-supervised segmentation, leveraging uncertainty estimation to refine pseudo labels. AdvChain enhances model robustness by incorporating adversarial data augmentation techniques including random noise, random bias field affine transformations, and diffeomorphic deformations. CTCT utilizes consistency regularization via co-training, improving learning stability. CLD was specifically designed to address the class imbalance issue in knee MRI segmentation. SCP-Net, a prototype-based approach, integrates both self-aware and cross-sample prototypes, improving performance by considering intra- and inter-class variations. DAE-MT introduces an anatomically-aware representation to estimate pixel-wise uncertainty from a single forward pass. Hybrid-Teacher combines 2D and 3D MT models with hybrid regularization, effectively capturing multi-dimensional features and enabling uncertainty-aware predictions. LeFeD utilizes feature-level discrepancies between dual decoders as feedback to the encoder, without relying on uncertainty modeling or strong regularization. Finally, PICK integrates masked image modeling into a three-decoder framework to jointly exploit pseudo label guidance, regional masking, and reconstruction for improved semi-supervised segmentation performance. Each of these methods offers a unique contribution to the field of semi-supervised segmentation, and we benchmark them against our proposed framework to evaluate its effectiveness. Tables 4 and 5 present the corresponding comparison results for two distinct datasets.

Table 4. Comparison of the proposed framework's performance with other semi-supervised segmentation methods on the OAI-iMorphics dataset.

| Method | DSC (%) | | | ASSD (mm) | | |
|---|---|---|---|---|---|---|
| | Lateral Meniscus | Medial Meniscus | Mean | Lateral Meniscus | Medial Meniscus | Mean |
| MT (2017) [31] | 86.04 | 81.90 | 83.97 | 0.41 | 0.98 | 0.70 |
| UA-MT (2019) [58] | 87.09 | 83.03 | 85.06 | 0.26 | 0.40 | 0.33 |
| AdvChain (2022) [59] | 88.31 | 83.61 | 85.96 | **0.19** | **0.37** | **0.28** |
| CTCT (2022) [60] | 86.44 | 81.74 | 84.09 | 0.44 | 0.65 | 0.55 |
| CLD (2022) [61] | 87.90 | 83.71 | 85.81 | 0.33 | **0.37** | 0.35 |
| SCP-Net (2023) [55] | 87.26 | 81.98 | 84.62 | 0.48 | 0.78 | 0.63 |
| DAE-MT (2024) [62] | 85.48 | 82.76 | 84.12 | 0.33 | 0.83 | 0.58 |
| Hybrid-Teacher (2024) [63] | 86.60 | 83.19 | 84.90 | 1.12 | 1.45 | 1.29 |
| LeFeD (2024) [46] | 88.16 | 83.32 | 85.74 | 0.24 | 0.44 | 0.34 |
| PICK (2025) [44] | 87.47 | 84.38 | 85.93 | 0.58 | 0.67 | 0.63 |
| ERANet (Ours) | **88.86** | **84.46** | **86.66** | 0.22 | 0.43 | 0.33 |

As Table 4 shows for the OAI-iMorphics dataset, our proposed method, ERANet, achieves the best performance with an average DSC of 86.66%. Compared with other state-of-the-art semi-supervised segmentation methods, ERANet demonstrates superior accuracy in segmenting both the lateral and medial meniscus, consistently outperforming existing methods in DSC. Among the other semi-supervised approaches, AdvChain, which incorporates adversarial data augmentation, outperforms regularization-based methods and demonstrates a slight improvement in ASSD. This reflects the importance of data augmentation when the model already has a certain level of segmentation capability. By jointly optimizing the data augmentation process and the segmentation network during training, challenging examples are generated, which enhances the model's generalizability for downstream tasks. However, implementing dynamic data augmentation requires significant computational resources. In contrast, our proposed framework leverages three crucial techniques (reviewed in Section 3) to effectively utilize limited labeled data and a substantial amount of unlabeled data: simulating meniscus variation through tailored data augmentation, aligning prototypical features, and selective ST. These strategies likely contribute to the superior performance of ERANet, even when only limited labeled scans are available. The representative segmentation results from different methods are presented in Figure 6.

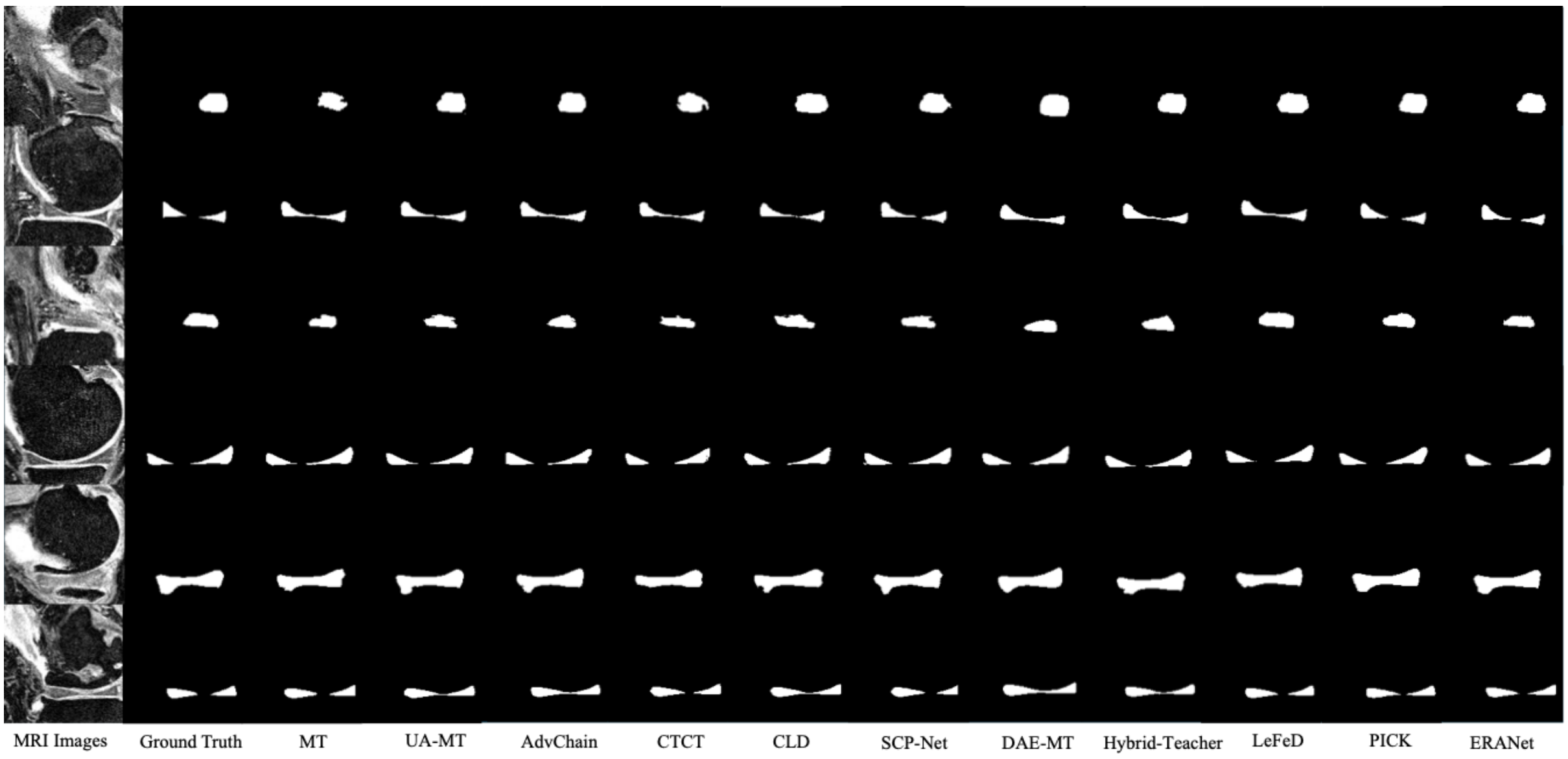

Figure 6. Segmentation results predicted by various state-of-the-art networks on the OAI-iMorphics dataset. From left to right: MRI Image, Ground Truth, and segmentation results from "MT," "UA-MT," "AdvChain," "CTCT," "CLD," "SCP-Net," "DAE-MT," "Hybrid-Teacher," "LeFeD," "PICK," and the proposed method.

The comparison results with other semi-supervised state-of-the-art methods for the in-house dataset are shown in Table 5. Similar to the OAI-iMorphics dataset, our proposed method, ERANet, achieves the best performance with an average DSC of 67.09%. The segmentation performance for the FSE sequence is lower than that for the DESS sequence, which can be attributed to the smaller amount of labeled data available for the FSE dataset, with only 5 labeled scans used for training. Additionally, the FSE dataset contains 36.4% of severe OA patients with a KL grade greater than 2. This introduces greater variability in the data, leading to higher discrepancies in the results from other methods. Compared with other semi-supervised methods, PICK demonstrates its superiority by incorporating a masked image modeling framework that leverages pseudo label guidance. This reflects the importance of guided learning from informative regions, as both PICK and our proposed CST adopt targeted supervision strategies. PICK enhances learning through mask-based reconstruction of pseudo-label-guided regions, while CST filters reliable predictions over time to provide consistent supervision when the model has already developed a certain level of segmentation capability. ERANet outperforms all models, likely due to the combination of data augmentation and feature alignment techniques that take full advantage of both labeled and unlabeled data. The representative segmentation results from different methods are presented in Figure 7.

Table 5. Comparison of the proposed framework's performance with other semi-supervised segmentation methods on the in-house dataset.

| Method | DSC (%) | | | ASSD (mm) | | |
|---|---|---|---|---|---|---|
| | Lateral Meniscus | Medial Meniscus | Mean | Lateral Meniscus | Medial Meniscus | Mean |
| MT (2017) [31] | 75.32 | 46.65 | 60.99 | 1.17 | 6.40 | 3.79 |
| UA-MT (2019) [58] | 76.37 | 47.72 | 62.05 | 0.93 | 5.54 | 3.24 |
| AdvChain (2022) [59] | 77.89 | 54.26 | 66.08 | 1.83 | 5.28 | 3.56 |
| CTCT (2022) [60] | 74.69 | 48.74 | 61.72 | 1.88 | 14.32 | 8.10 |
| CLD (2022) [61] | 77.17 | 53.13 | 65.15 | 1.98 | **3.91** | 2.95 |
| SCP-Net (2023) [55] | 76.56 | 48.28 | 62.42 | **0.65** | 4.57 | **2.61** |
| DAE-MT (2024) [62] | 78.40 | 48.24 | 63.32 | 1.29 | 12.51 | 6.9 |
| Hybrid-Teacher (2024) [63] | 77.38 | 52.45 | 64.92 | 2.07 | 7.18 | 4.63 |
| LeFeD (2024) [46] | 76.68 | 52.84 | 64.76 | 2.52 | 10.89 | 6.71 |
| PICK (2025) [44] | 77.80 | 55.25 | 66.53 | 1.05 | 4.24 | 2.65 |
| ERANet (Ours) | **78.17** | **56.01** | **67.09** | 1.06 | 5.66 | 3.36 |

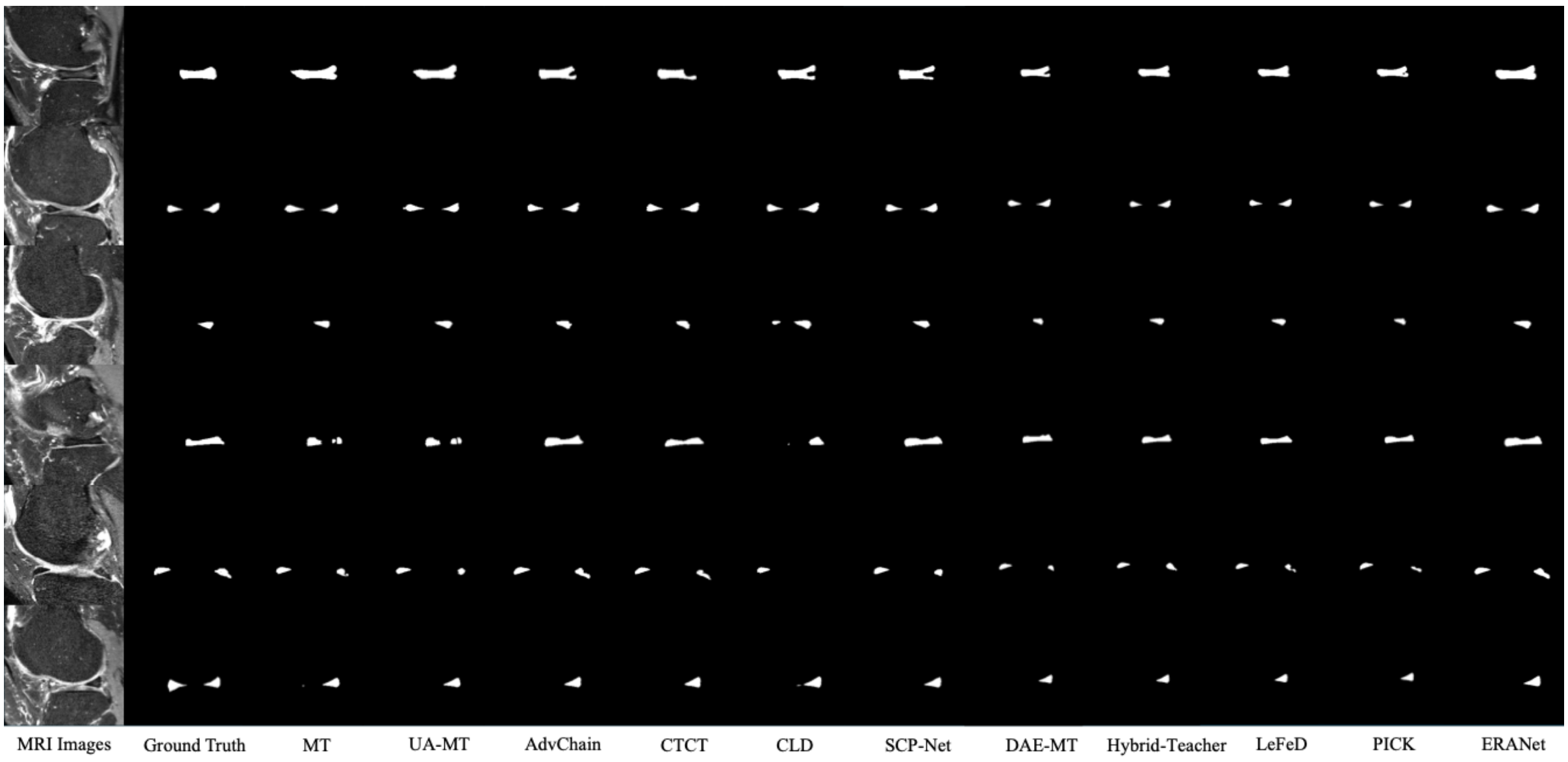

Figure 7. Segmentation results predicted by various state-of-the-art networks on the in-house dataset. From left to right: MRI Image, Ground Truth, and segmentation results from "MT," "UA-MT," "AdvChain," "CTCT," "CLD," "SCP-Net," "DAE-MT," "Hybrid-Teacher," "LeFeD," "PICK," and the proposed method.

Compared with recent state-of-the-art semi-supervised segmentation methods, ERANet improves performance through complementary advances at the data and learning levels. ERA enhances robustness by introducing anatomically boundary variations instead of the adversarial perturbations used in AdvChain [59]. This targeted strategy improves boundary delineation and better preserves structural integrity in small-structure segmentation. PCA promotes compact and separable feature clusters, improving feature stability and alignment beyond feature-level discrepancy or reconstruction-based regularization methods such as LeFeD [46] and PICK [44]. CST further improves pseudo-label reliability by enforcing temporal consistency across checkpoints and achieves greater stability than MT- or co-training–based [58, 60 , 62, 63] frameworks that rely on single-step consistency. These components together improve segmentation accuracy under limited supervision, although PCA and CST increase training complexity.

## 4.6 Ablation Study and Component Analysis

### 4.6.1. Ablation Study

To evaluate the individual and combined contributions of the proposed modules: ERA, PCA, and CST, we conducted an ablation study based on the basic MT structure using the OAI-iMorphics dataset. The results are summarized in Table 6. The baseline MT model achieved a mean DSC of 83.97%. Introducing each module individually improved both metrics, with ERA providing the largest improvement in lateral segmentation (DSC: +1.41%), and CST yielding the greatest gain in medial accuracy (DSC: +1.16%), likely due to its ability to filter out noisy pseudo labels. PCA improved feature consistency and yielded a mean DSC of 84.81%. Combining two modules further enhanced performance. The ERA+PCA setting improved spatial boundary delineation (mean ASSD: 0.38 mm), while ERA+CST achieved the highest mean DSC among dual combinations (85.79%) and strong boundary localization (mean ASSD: 0.35 mm). The PCA+CST configuration balanced feature compactness and pseudo label stability, resulting in competitive segmentation quality. The full ERANet model, integrating all three modules, achieved the best performance with a mean DSC of 86.66% and the lowest mean ASSD of 0.33 mm, demonstrating the complementary effects of the proposed modules and confirming their synergy.

Table 6. Ablation study of ERANet showing the impact of ERA, PCA, and CST modules on segmentation performance (DSC and ASSD) on the OAI-iMorphics dataset.

| Method | ERA | PCA | CST | DSC (%) | | | ASSD (mm) | | |
|---|---|---|---|---|---|---|---|---|---|
| | | | | Lateral Meniscus | Medial Meniscus | Mean | Lateral Meniscus | Medial Meniscus | Mean |
| MT | ✗ | ✗ | ✗ | 86.04 | 81.90 | 83.97 | 0.41 | 0.98 | 0.70 |
| Only ERA | ✓ | ✗ | ✗ | 87.45 | 82.81 | 85.13 | 0.34 | 0.59 | 0.47 |
| Only PCA | ✗ | ✓ | ✗ | 86.67 | 82.95 | 84.81 | 0.31 | 0.55 | 0.43 |
| Only CST | ✗ | ✗ | ✓ | 87.11 | 83.06 | 85.09 | 0.29 | 0.71 | 0.50 |
| ERA + PCA | ✓ | ✓ | ✗ | 88.01 | 82.94 | 85.48 | 0.22 | 0.54 | 0.38 |
| ERA + CST | ✓ | ✗ | ✓ | 87.99 | 83.58 | 85.79 | 0.25 | 0.45 | 0.35 |
| PCA + CST | ✗ | ✓ | ✓ | 87.72 | 83.47 | 85.60 | 0.31 | 0.47 | 0.39 |
| ERANet (Ours) | ✓ | ✓ | ✓ | **88.86** | **84.46** | **86.66** | **0.22** | **0.43** | **0.33** |

Figure 8 presents the feature activation maps along with ground truth annotations and ERANet predictions for four representative MRI slices, comparing different module configurations within the ERANet framework. For each method, the visualized maps highlight the most relevant feature channel from the final decoder layer, offering insights into model attention and localization of meniscal structures.

The baseline MT model yields diffused activations with limited localization to the meniscal regions. Incorporating ERA alone enhances boundary variance and activation sharpness, reflecting improved sensitivity to anatomical structures. PCA alone produces smoother but less discriminative feature maps, indicating stronger intra-class compactness but weaker boundary precision. CST alone suppresses spurious responses in background regions and sharpens focus on the meniscus, likely due to its ability to filter noisy pseudo labels. When combining modules, their effects become complementary. The ERA+PCA configuration demonstrates both edge sensitivity and compact feature activation. ERA+CST further suppresses irrelevant regions while enhancing anatomically specific responses. PCA+CST produces well-separated activations across meniscal zones, reflecting stable pseudo label propagation and clearer class-wise representation. The full ERANet model, integrating all three modules, achieves the most focused and precise activations, with responses highly localized to the ground-truth meniscus boundaries. The corresponding predictions show the closest alignment to manual annotations, confirming the synergistic effectiveness of combining structural augmentation, feature alignment, and selective ST.

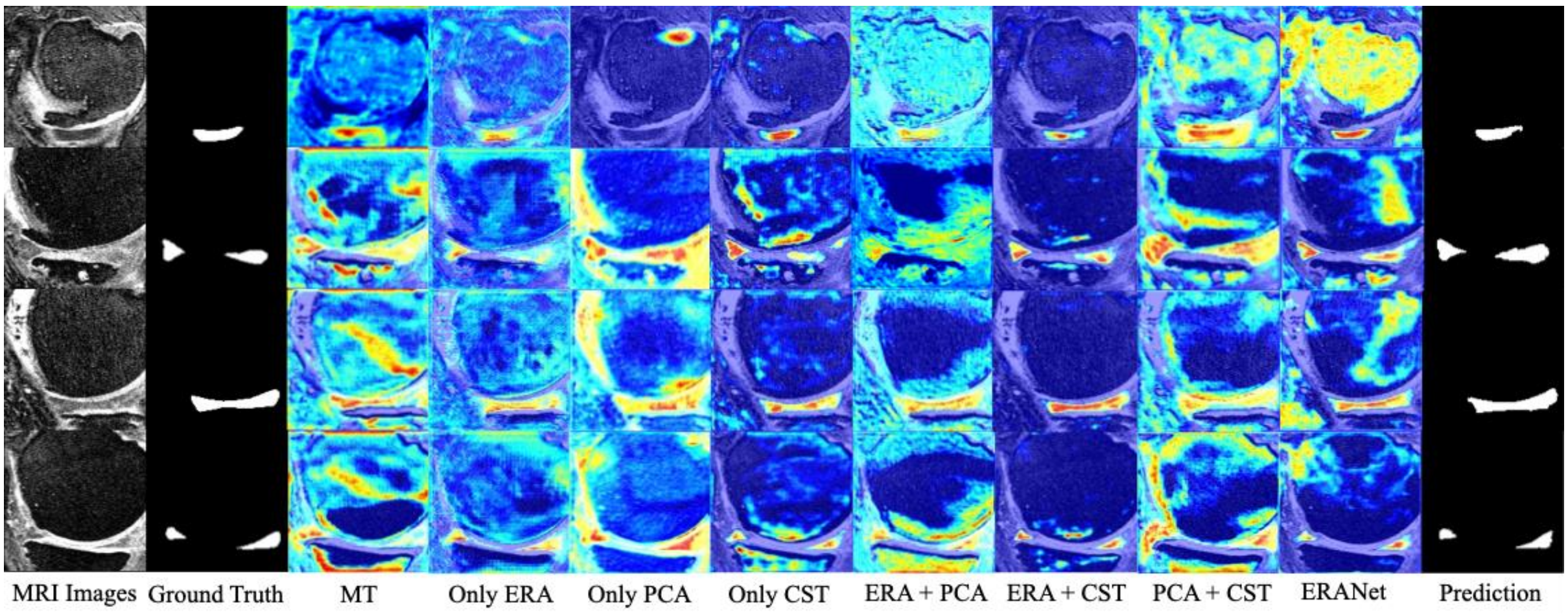

Figure 8. Visualization of feature activation maps across different ablation settings of ERANet and segmentation predictions. The highlighted channel from the final decoder layer reflects the network's attention to meniscal structures.

### 4.6.2. Component Analysis

#### i) Edge Replacement Augmentation

As illustrated in Figures 9 and 10, ERA is integrated into the U1 and U3 networks to enhance segmentation of target tissues. Prototypical consistency loss and CST are applied as self-supervision techniques to refine the predictions of both models. Fig. 9 presents the performance of the U1 model under various augmentation strategies, including the absence of augmentation ("W/O Aug") and the use of alternative advanced methods. "Strong Aug" encompasses random augmentations like Gaussian blur and color jittering, while advanced methods incorporate information from multiple images. For instance, "Mixup" creates new samples by linearly blending two examples and their labels, while variations such as CutMix replace rectangular regions of images, and Copy-Paste transfers specific object pixels. Similarly, Fig. 10 examines the U3 model's performance under the same augmentation settings. The integration of ERA significantly improves the segmentation accuracy of both U1 (Fig. 9) and U3 (Fig. 10), yielding higher average DSC scores, especially for the medial meniscus. In Fig. 10, the baseline U3 model refers to the model trained using all reliable unlabeled data treated as pseudo-labeled data. This improvement can be attributed to the medial meniscus being more prone to injury compared to the lateral meniscus, with ERA introducing morphological variations that better mimic its structural diversity, thereby enhancing segmentation performance.

To further validate the effectiveness of ERA within ERANet, we assessed its impact on segmentation performance under varying ratios of annotated to unannotated data. The baseline U1 and U3 models were trained with and without the inclusion of ERA. As shown in Fig. 11, the semi-supervised framework consistently demonstrated improved performance with the proposed ERA, regardless of the proportion of annotated to unannotated data.

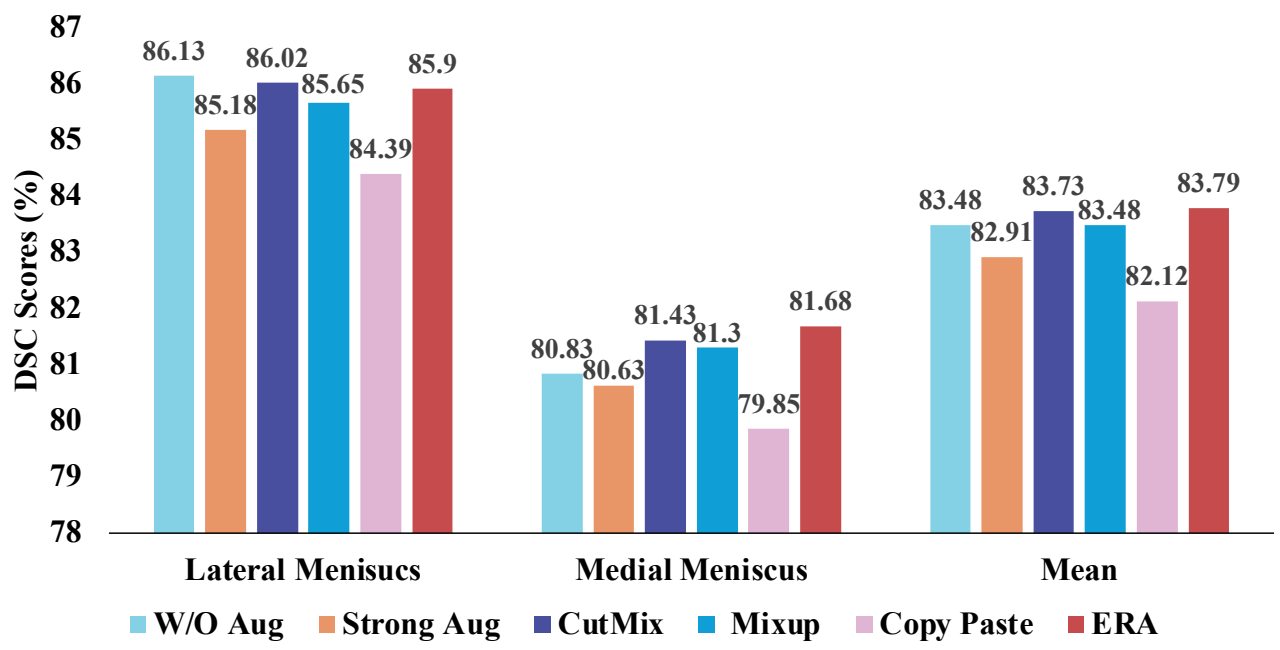

Figure 9. Quantitative evaluation of the baseline U1 model under various augmentation strategies. "Mean" refers to the average value calculated from the lateral meniscus and medial meniscus.

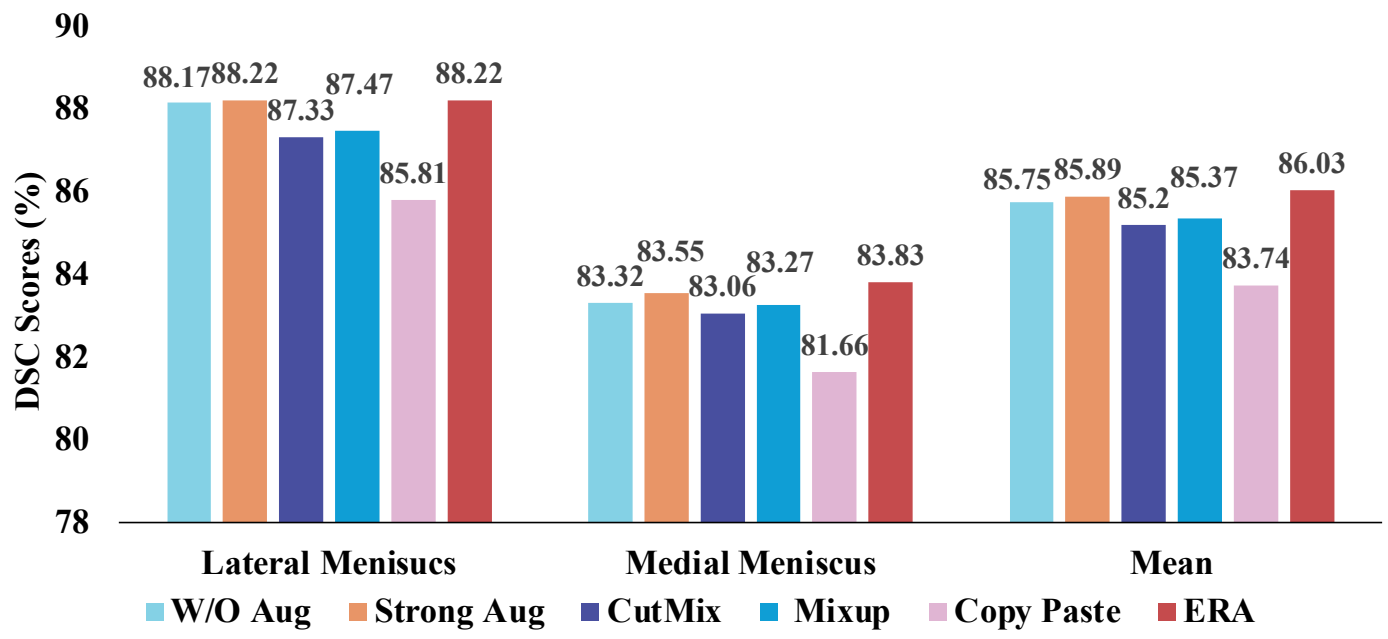

Figure 10. Quantitative evaluation of the baseline U3 model under various augmentation strategies. "Mean" refers to the average value calculated from the lateral meniscus and medial meniscus.

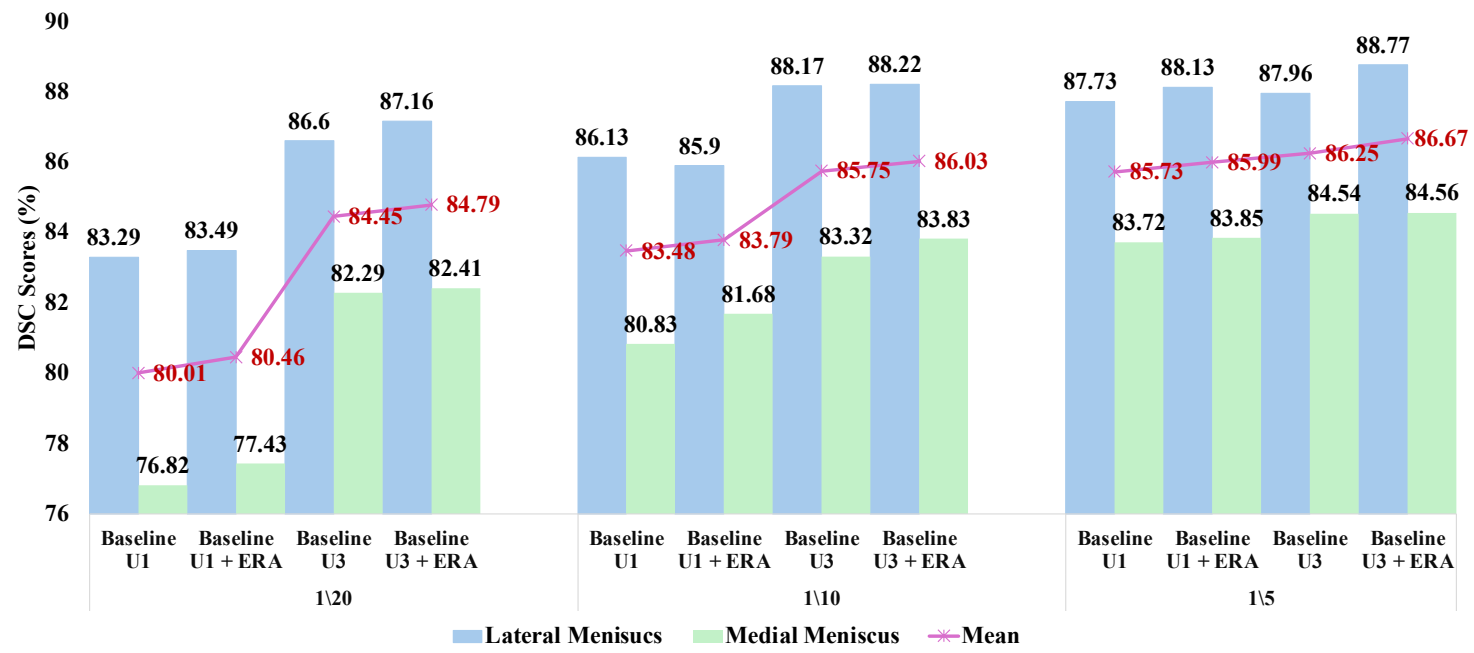

Figure 11. Impact of ERA on segmentation performance with varying ratios of annotated to unannotated data. "Mean" refers to the average value calculated from the lateral meniscus and medial meniscus.

### ii) Prototype Consistency Alignment

Building upon existing experiments, we evaluate the effectiveness of PCA by comparing the baseline U1 model trained with and without the PCA module. In the proposed framework, ERA has already been validated as a means to enhance meniscus segmenation performance through data perturbation. Therefore, we incorporate PCA on unlabeled images, building upon the foundation established by ERA. Note that the baseline U1 model, when equipped with ERA and PCA, corresponds to the U1 model presented in Table 2. Fig. 12 presents the comparison results under varying ratios of annotated to unannotated data. As shown in Fig. 12, the semi-supervised framework consistently demonstrates improved performance with the proposed PCA, regardless of the ratio of annotated to unannotated data. Notably, the benefit of PCA becomes more pronounced as the number of labeled images decreases, highlighting its effectiveness in scenarios with limited annotated data.

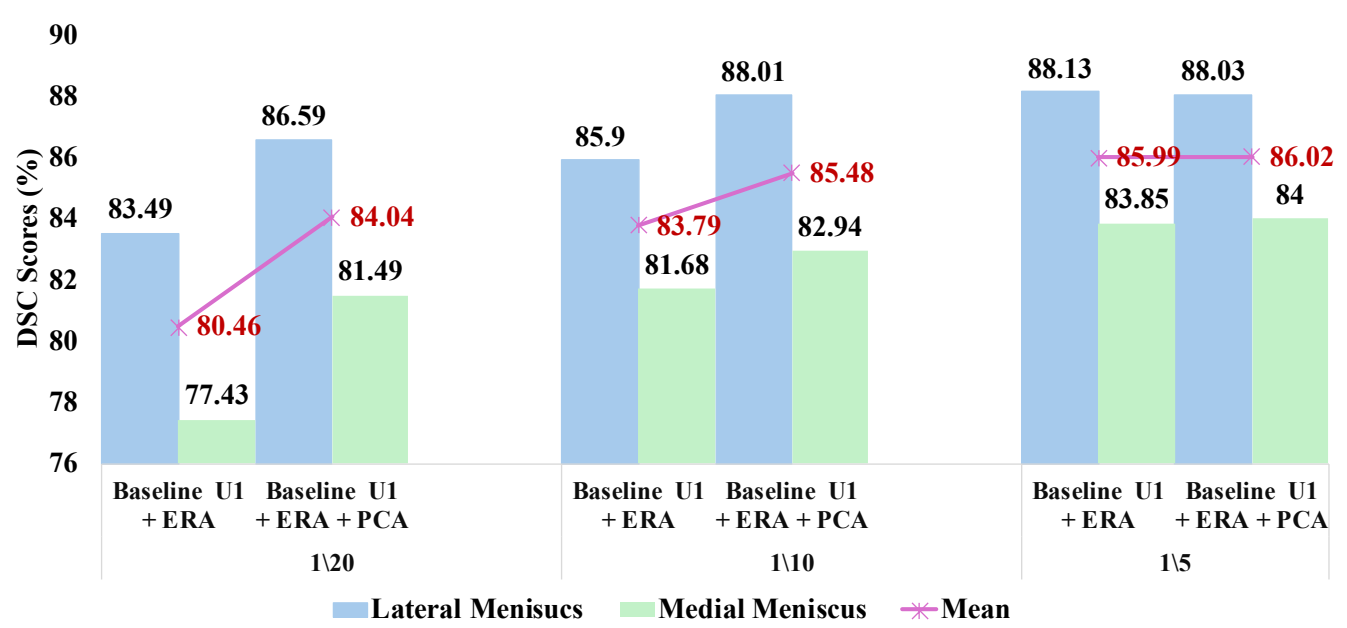

Figure 12. Comparison of segmentation performance for the U1 model with and without the prototype consistency alignment (PCA) module under varying ratios of annotated to unannotated data. "Mean" refers to the average value calculated from the lateral meniscus and medial meniscus.

### iii) Conditional Self-Training Strategy

To validate the effectiveness of CST within ERANet, we evaluated the impact of the two-stage iterative CST strategy on segmentation performance. The model is progressively optimized using a combination of manually labeled and pseudo-labeled images. As reported in Table 7, the semi-supervised framework consistently benefits from the proposed CST, irrespective of the ratio of annotated to unannotated data, highlighting CST's robustness and superior segmentation accuracy across varying annotation levels. Note that integrating CST into the baseline U3 model equipped with ERA results in the final proposed U4 model. Further analysis, detailed in Table 8, examines the improvements achieved at each stage of the CST process. During the first stage, where only a subset of the unlabeled images is used, the segmentation results surpass those of U1, underscoring the reliability of the selected pseudo labels. Ablation studies further reveal that the default DSC score threshold of 80% effectively balances performance, with ERANet demonstrating robustness to threshold variations. Lower thresholds included more uncertain pseudo-labels and introduced noise that reduced accuracy. Higher thresholds retained too few samples to provide sufficient supervision. As shown in Table 8, a threshold of 0.8 achieved a good balance between label reliability and data utilization.

Table 7. Quantitative comparison of segmentation performance between common ST and CST on the OAI-Imorphics dataset with different data partition protocols. The baseline U3 model equipped with ERA and CST is equivalent to the proposed ERANet.

| | #scans used | | DSC (%) | | | ASSD (mm) | | |
|---|---|---|---|---|---|---|---|---|
| | Annotated | Unannotated | Lateral Meniscus | Medial Meniscus | Mean | Lateral Meniscus | Medial Meniscus | Mean |
| Baseline U3 + ERA | 28 | 112 | 88.77 | 84.56 | 86.67 | 0.25 | 0.38 | 0.32 |
| Baseline U3+ERA+CST | 28 | 112 | **88.98** | **85.15** | **87.07** | **0.22** | 0.41 | **0.32** |
| Baseline U3 + ERA | 14 | 126 | 88.22 | 83.83 | 86.03 | 0.27 | 0.37 | 0.32 |
| Baseline U3+ERA+CST | 14 | 126 | **88.86** | **84.46** | **86.66** | **0.22** | 0.43 | 0.33 |
| Baseline U3 + ERA | 7 | 133 | 87.16 | 82.41 | 84.79 | 0.30 | 0.51 | 0.41 |
| Baseline U3+ERA+CST | 7 | 133 | **87.52** | **83.28** | **85.40** | **0.22** | **0.35** | **0.29** |

Table 8. Training process of the proposed CST. CST initially trains the model on reliable unlabeled images (Self-training #1) and then on all pseudo-labeled images (Self-training #2). A quantitative comparison of segmentation performance during the CST training process on the OAI-Imorphics dataset with different data partition protocols.

| DSC threshold of reliable images | Stage | DSC (%) | | | ASSD (mm) | | |
|---|---|---|---|---|---|---|---|
| | | Lateral Meniscus | Medial Meniscus | Mean | Lateral Meniscus | Medial Meniscus | Mean |
| 50% | Self-training #1 | 87.96 | 83.65 | 85.81 | 0.31 | 0.45 | 0.38 |
| | Self-training #2 | **88.38** | **84.10** | **86.24** | **0.25** | 0.51 | **0.38** |
| 60% | Self-training #1 | 88.10 | 83.88 | 85.99 | 0.26 | 0.38 | 0.32 |
| | Self-training #2 | **88.29** | **84.23** | **86.26** | **0.22** | **0.38** | **0.30** |
| 80% (default) | Self-training #1 | 88.34 | 83.88 | 86.11 | 0.25 | 0.42 | 0.34 |
| | Self-training #2 | **88.86** | **84.46** | **86.66** | **0.22** | 0.43 | **0.33** |
| 100% | Self-training #1 | 88.22 | 83.83 | 86.03 | 0.27 | 0.37 | 0.32 |
| | Self-training #2 | **88.29** | **84.62** | **86.46** | **0.27** | 0.43 | 0.35 |

## 4.7 Hyperparameter Sensitive Analysis

### 4.7.1. Sensitivity to Loss Function Weights

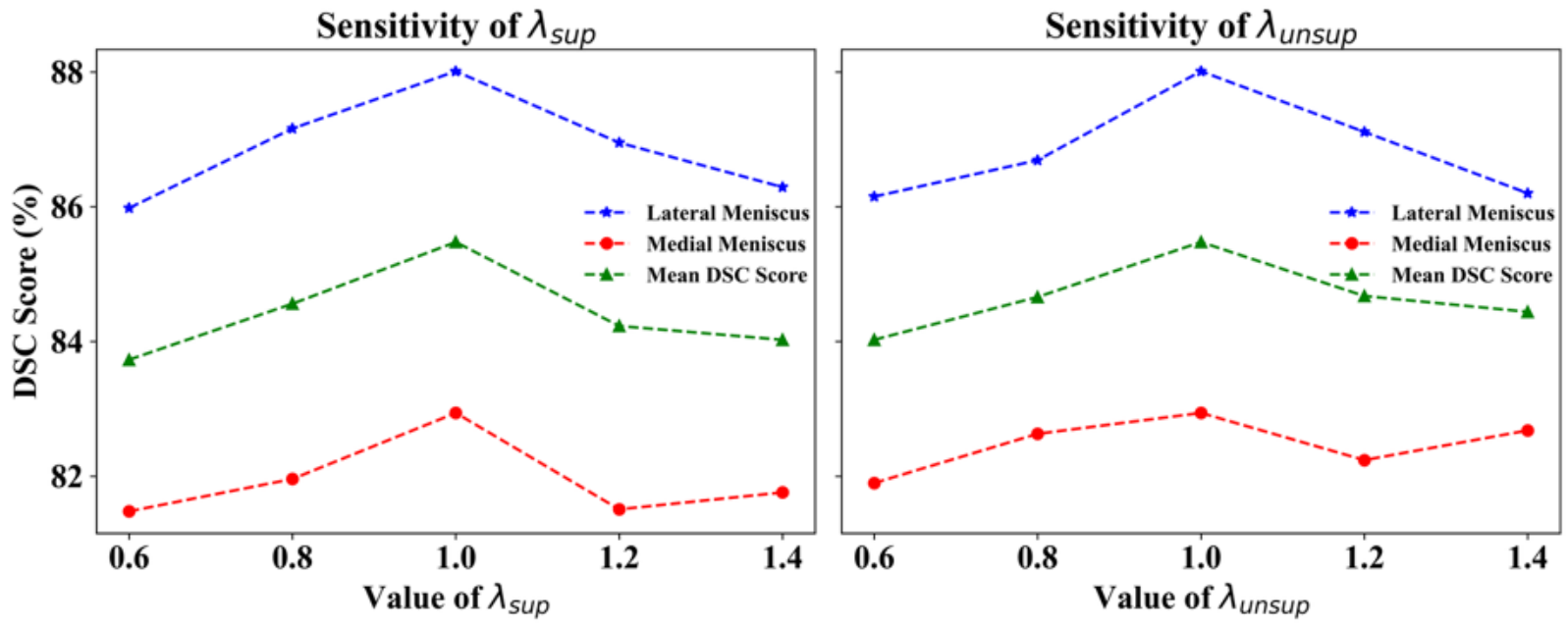

Figure 13. Sensitivity analysis of $\lambda_{sup}$ and $\lambda_{unsup}$ in ERANet.

The hyperparameters $\lambda_{sup}$ and $\lambda_{unsup}$ control the weighting between the supervised segmentation loss and the unsupervised cross-entropy loss in our ERANet framework, specifically within the U1 model. To evaluate their influence, we performed a sensitivity analysis by varying each parameter independently within the range [0.6,1.4] in increments of 0.2, while fixing the other at 1.0. The results, summarized in Fig. 13, reveal a bell-shaped performance trend for both parameters. Optimal segmentation performance was achieved when setting $\lambda_{sup}$ = 1.0 and $\lambda_{unsup}$=1.0, yielding the highest DSC scores. Performance declined slightly at both lower and higher values, suggesting that either under- or over-weighting the corresponding loss term may impair generalization. Additionally, the lateral meniscus was found to be more sensitive to $\lambda_{sup}$, whereas the medial meniscus benefited more from careful tuning of $\lambda_{unsup}$, likely due to greater anatomical variability in the medial region.

### 4.7.2. Sensitivity to the ERA Replacement Strategy

In the proposed ERA, the edge-replacement ratio $r_e$ controls the width of the replaced region along the outer meniscal margin. As shown in Table 9, increasing $r_e$ from 0.1 to 0.3 raised the mean DSC from 85.98% to 86.66%. When $r_e$ increased to 0.4, the mean DSC slightly dropped to 86.28%, which means that too large a replacement range may affect anatomical consistency. A moderate value of $r_e = 0.3$ provides the best balance between structural integrity and augmentation diversity.

Table 9. Sensitivity analysis of the edge-replacement ratio $r_e$ in ERA.

| $r_e$ | DSC (%) | | |
|---|---|---|---|
| | Lateral Meniscus | Medial Meniscus | Mean |
| 0.1 | 87.95 | 84.01 | 85.98 |
| 0.2 | 88.23 | 84.34 | 86.29 |
| **0.3** | **88.86** | **84.46** | **86.66** |
| 0.4 | 88.13 | 84.42 | 86.28 |

### 4.7.3. Sensitivity to Checkpoint Design in CST

In CST, pseudo-label reliability was evaluated from their temporal stability across $K$ checkpoints saved during training. By default, five evenly spaced checkpoints were used (at 1/5–5/5 of the total iterations. To examine the influence of checkpoint design, we varied the number of checkpoints ($K \in \{3,5,7\}$) and their temporal placement under two schedules: (i) uniform, with evenly spaced checkpoints, and (ii) back-loaded, with checkpoints concentrated near the end of training. Specifically, the back-loaded checkpoints were located at epochs [140, 180, 200] for $K = 3$; [100, 140, 170, 190, 200] for $K = 5$; and [100, 130, 156, 176, 188, 196, 200] for $K = 7$. As summarized in Table 10, increasing $K$ from 3 to 5 under the uniform schedule improved the mean DSC from 86.14% to 86.66%, whereas performance plateaued at $K = 7$. The back-loaded schedule consistently produced slightly lower results (85.88%, 86.27%, and 86.45% for $K = 3, 5, 7$, respectively), which may result from the high correlation among late-stage checkpoints and the limited temporal diversity of their pseudo-labels. Larger $K$ values yielded accuracy gains. Reducing $K$ to 3 lowered stability and mean DSC. Overall, a configuration of $K = 5$ with a uniform schedule provided the best balance between computational efficiency and segmentation accuracy.

Table 10. Sensitivity analysis of checkpoint design in CST.

| K | Schedule | DSC (%) | | |
|---|---|---|---|---|
| | | Lateral Meniscus | Medial Meniscus | Mean |
| 3 | Uniform | 88.31 | 83.96 | 86.14 |
| | Back-loaded | 88.25 | 83.51 | 85.88 |
| 5 | Uniform | 88.86 | 84.46 | 86.66 |
| | Back-loaded | 88.53 | 84.01 | 86.27 |
| 7 | Uniform | 88.89 | 84.61 | 86.75 |
| | Back-loaded | 88.55 | 84.35 | 86.45 |

The hyperparameter analysis demonstrates how key design choices influence our model stability and generalization. Balanced loss weights ($\lambda_{sup} = \lambda_{unsup} = 1.0$) maintain effective supervised guidance while enabling the extraction of informative structural features from unlabeled data. A moderate edge-replacement ratio ($r_e = 0.3$) produces anatomically plausible augmentation without disturbing structural integrity. For small structures segmentation, augmenting potential lesion areas while preserving normal anatomy yields the best balance. In CST, using five uniformly spaced checkpoints (K=5) ensures reliable pseudo-label stability with low computational cost. This design may enhance the reliability of semi-supervised learning in longitudinal or multi-site MRI studies, where unlabeled data are abundant but annotation consistency is limited. These settings collectively support consistent performance in real-world MRI applications and make the ERANet practical for scalable clinical deployment.

## 5. Limitations and Future Work

While ERANet demonstrates strong performance, several limitations remain. ERA is specifically tailored to meniscus morphology and may not generalize well to other anatomical structures. PCA depends on the quality of pseudo labels, which can be affected by early-stage prediction noise. CST requires multi-checkpoint stability analysis and fixed thresholds, introducing computational overhead and potentially overlooking samples that improve later in training. Moreover, due to the limited availability of annotated meniscus datasets, the method was validated only on two MRI sequences (DESS and FSE) with differing contrast characteristics, which may constrain generalizability. Future work will streamline the training pipeline using adaptive reliability metrics and parallel self-training, and extend the framework to automatic meniscal quantification. In this regard, multi-modality fusion and emerging spiking neural models hold promise for improving generalizability and reducing computational cost [64, 65]. Furthermore, recent studies [66] have demonstrated the potential of graph-based representations in capturing structural features of the knee joints and improving OA diagnosis accuracy.

## 6. Conclusion

In this study, we propose an effective semi-supervised framework that integrates ERA, PCA, and CST to enhance meniscus segmentation performance in knee MRI images. The framework utilizes both manually labeled and pseudo-labeled data, integrating ERA, prototypical consistency loss and CST employed as self-supervision strategies to refine model predictions. Extensive experiments were conducted on the OAI-Imorphics dataset and the in-house dataset, evaluating the proposed method's effectiveness across varying ratios of annotated to unannotated data. Our results show that the combination of ERA, prototype consistency alignment, and CST consistently improves segmentation accuracy, particularly for the medial meniscus.

**Declaration of competing interest**

The authors declare that there are no conflicts of interest regarding the publication of this paper.

**CRediT authorship contribution statement**

**Siyue Li**: Conceptualization; Data Curation; Formal Analysis; Investigation; Methodology; Software; Validation; Visualization; Writing-Original Draft; Writing - Review & Editing. **Yongcheng Yao**: Software; Validation; Investigation; Writing - Review & Editing. **Junru Zhong**: Validation; Data Curation; Investigation; Writing - Review & Editing. **Shutian Zhao**: Investigation; Data Curation; Writing - Review & Editing. **Fan Xiao**: Data Curation; Writing - Review & Editing. **Tim-Yun Michael Ong**: Data Curation; Resources; Writing - Review & Editing. **Ki-Wai Kevin Ho**: Data Curation; Resources; Writing - Review & Editing. **James F. Griffith**: Data Curation; Resources; Writing - Review & Editing. **Yudong Zhang**: Validation, Writing-Review & Editing. **Shuiua, Wang**: Validation, Writing Review & Editing. **Jin Hong**: Conceptualization, Investigation; Methodology; Funding Acquisition; Supervision; Writing-Review & Editing. **Weitian Chen**: Investigation; Methodology, Resources; Funding Acquisition; Project Administration; Supervision; Writing-Review & Editing.

**Acknowledgments**

This study was supported by in part by the National Natural Science Foundation of China (62466033), in part by the Jiangxi Provincial Natural Science Foundation (20242BAB20070) and a grant from the Innovation and Technology Commission of the Hong Kong SAR (Project No. MRP/001/18X). We would like to acknowledge Chi Yin Ben Choi and Cheuk Nam Cherry Cheng for their assistance in patient recruitment and MRI exams and Tsz Shing Adam Kwong for his assistance in data processing.